%% file: arxiv_icml.tex
\documentclass{article}

\usepackage{microtype}
\usepackage{graphicx}
\usepackage{subfigure}
\usepackage{booktabs} 

\usepackage{hyperref}

\usepackage[accepted]{icml2023}

\usepackage{amsmath}
\usepackage{amssymb}
\usepackage{mathtools}
\usepackage{amsthm}
\usepackage{makecell}

\usepackage[capitalize,noabbrev]{cleveref}

\theoremstyle{plain}
\newtheorem{theorem}{Theorem}[section]
\newtheorem{proposition}[theorem]{Proposition}
\newtheorem{lemma}[theorem]{Lemma}

\theoremstyle{definition}
\newtheorem{definition}[theorem]{Definition}

\theoremstyle{remark}

\usepackage[textsize=tiny]{todonotes}

\input{math_commands.tex}

\usepackage[utf8]{inputenc}
\usepackage[T1]{fontenc}    
\usepackage{graphicx} 
\usepackage{subfigure}
\usepackage{wrapfig}
\usepackage{xspace}
\usepackage{caption}
\usepackage{booktabs}
\usepackage{amssymb}
\usepackage{wrapfig}
\usepackage{tikz}
\usetikzlibrary{arrows,automata}
\tikzset{terminal state/.style={draw,rectangle,minimum size=.3in}}

\usepackage[noend]{algpseudocode}
\usepackage{subfigure}
\usepackage{multirow}

\usepackage{amsmath}

\usepackage{enumitem}

\newcommand{\sd}{s}
\newcommand{\sx}{e}

\newcommand{\Sf}{\mathcal{Z}}

\newcommand{\sff}{z}

\usepackage{amsthm}

\usepackage{pifont}
\usepackage{multirow}
\usepackage{graphicx}

\definecolor{myblue}{RGB}{43, 115, 219}

\newcommand{\cmark}{\ding{51}}%
\newcommand{\bluecheck}{{\color{myblue}\cmark}}
\newcommand{\xmark}{\ding{55}}%

\newcommand{\tablestyle}[2]{\setlength{\tabcolsep}{#1}\renewcommand{\arraystretch}{#2}\centering\footnotesize}
\newlength\savewidth\newcommand\shline{\noalign{\global\savewidth\arrayrulewidth
  \global\arrayrulewidth 1pt}\hline\noalign{\global\arrayrulewidth\savewidth}}

\usepackage{ulem}

\usepackage{hyperref}
\usepackage{url}
\usepackage{std_definitions}

\newcommand{\supp}{{\tt supp}}

\newcommand{\methodname}{\textsc{ACRO}\xspace}
\newcommand{\easyexo}{\textsc{Easy-Exo}\xspace}
\newcommand{\mediumexo}{\textsc{Medium-Exo}\xspace}
\newcommand{\hardexo}{\textsc{Hard-Exo}\xspace}

\newcommand{\exodatasets}{\textsc{Exogenous Datasets}\xspace}
\newcommand{\highlight}[1]{\colorbox{blue!10}{#1}}
\newcommand{\redlight}[1]{\colorbox{red!10}{#1}}
\definecolor{mygray}{gray}{0.4}

\newcommand{\g}[2]{#1\textsubscript{\textcolor{mygray}{$\pm$#2}}}
\icmltitlerunning{Principled Offline RL in the Presence of Rich Exogenous Information}

\begin{document}

\twocolumn[
\icmltitle{Principled Offline RL in the Presence of Rich Exogenous Information}



\icmlsetsymbol{equal}{*}

\begin{icmlauthorlist}
\icmlauthor{Riashat Islam}{equal,Mcgill,MSR Montreal,MSR NYC}
\icmlauthor{Manan Tomar}{equal,UAlberta,MSR Montreal}
\icmlauthor{Alex Lamb}{MSR NYC}
\icmlauthor{Yonathan Efroni}{Meta}
\icmlauthor{Hongyu Zang}{Beijing}
\icmlauthor{Aniket Didolkar}{Udem,MSR NYC}
\icmlauthor{Dipendra Misra}{MSR NYC}
\icmlauthor{Xin Li}{Beijing}
\icmlauthor{Harm Van Seijen}{MSR Montreal}
\icmlauthor{Remi Tachet Des Combes}{MSR Montreal}
\icmlauthor{John Langford}{MSR NYC}
\end{icmlauthorlist}

\author{\parbox{\linewidth}{Riashat Islam$^{*1,6,7}$, Manan Tomar\thanks{Equal Contribution; Corresponding Author E-mails: riashat.islam@mail.mcgill.ca, manantomar@gmail.com, lambalex@microsoft.com, jcl@microsoft.com}~~$^{2,6}$, Alex Lamb$^7$, Yonathan Efroni$^{5}$,
Hongyu Zang$^{4}$, \\ Aniket Didolkar$^{3, 7}$, Dipendra Misra$^{7}$, Xin Li$^{4}$
Harm Van Seijen$^{6}$, \\ Remi Tachet Des Combes$^{6}$, John Langford$^{7}$} \vspace{0.3em} \\ 
$^1$ McGill University, Quebec AI Institute \quad
$^2$ University of Alberta, Amii\\
$^3$ University of Montreal, Quebec AI Institute \quad
$^4$ Beijing Institute of Technology, Beijing\\ 
$^5$ Meta, New York\quad
$^6$ Microsoft Research, Montreal\quad
$^7$ Microsoft Research, New York\\
}

\icmlaffiliation{Mcgill}{McGill University, Quebec AI Institute}
\icmlaffiliation{MSR Montreal}{Microsoft Research, Montreal}
\icmlaffiliation{MSR NYC}{Microsoft Research, New York}
\icmlaffiliation{UAlberta}{University of Alberta}
\icmlaffiliation{Meta}{Meta, New York}
\icmlaffiliation{Beijing}{Beijing Institute of Technology, Beijing}
\icmlaffiliation{Udem}{University of Montreal, Quebec AI Institute}

\icmlcorrespondingauthor{Riashat Islam}{riashat.islam@mail.mcgill.ca}
\icmlcorrespondingauthor{Manan Tomar}{manan.tomar@gmail.com}
\icmlcorrespondingauthor{Alex Lamb}{lambalex@microsoft.com}
\icmlcorrespondingauthor{John Langford}{jcl@microsoft.com}

\icmlkeywords{Machine Learning, ICML}

\vskip 0.3in
]

\printAffiliationsAndNotice{\icmlEqualContribution} 

\begin{abstract}
Learning to control an agent from offline data collected in a rich pixel-based visual observation space is vital for real-world applications of reinforcement learning (RL). A major challenge in this setting is the presence of input information that is hard to model and irrelevant to controlling the agent. This problem has been approached by the theoretical RL community through the lens of \textit{exogenous information}, i.e., any control-irrelevant information contained in observations. For example, a robot navigating in busy streets needs to ignore irrelevant information, such as other people walking in the background, textures of objects, or birds in the sky. In this paper, we focus on the setting with visually detailed exogenous information and introduce \href{https://drive.google.com/drive/folders/1HsksquQ6gKQUDj_1Qe7dc-R_h8m6A1H3?usp=drive_link}{new offline RL benchmarks} that offer the ability to study this problem. We find that contemporary representation learning techniques can fail on datasets where the noise is a complex and time-dependent process, which is prevalent in practical applications. To address these, we propose to use multi-step inverse models to learn Agent-Centric Representations for Offline-RL (\methodname). Despite being simple and reward-free, we show theoretically and empirically that the representation created by this objective greatly outperforms baselines. Code is provided at \url{https://github.com/manantomar/agent-centric-representations}.
\end{abstract}

\section{Introduction}
\vspace{-1mm}
Effective real-world applications of reinforcement learning or sequential decision-making must cope with exogenous information in sensory data.  For example, visual datasets of a robot or car navigating in busy city streets might contain information such as advertisement billboards, birds in the sky, or other people crossing the road.   Parts of the observation (such as birds in the sky) are irrelevant for controlling the agent, while other parts (such as people crossing along the navigation route) are extremely relevant.  How can we effectively learn a representation of the world that extracts just the relevant information for controlling the agent while ignoring irrelevant information?

Real-world tasks are often more easily solved with fixed offline datasets since operating from offline data enables thorough testing before deployment, which can ensure safety, reliability, and quality in the deployed policy \citep{ Lange2012BatchRL,offlineRobotics, KumarFSTL19, offlineDialogue, OfflineTutorial}.  The Offline-RL setting also eliminates the need to address exploration and planning, which come into play during data collection.\footnote{This elimination, however, can make offline RL more difficult if the wrong data is collected.}  Although approaches from representation learning have been studied in the online case, yielding improvements, exogenous information has proved to be empirically challenging. In this paper, we therefore ask the question: is it possible to learn distraction-invariant representations from rich observations in offline RL?

Approaches for discovering small tabular MDPs (${\leq}500$ discrete latent states) or linear control problems invariant to exogenous information have been introduced \citep{dietterich2018discovering,efroni2021provably,exoRL,efroni2022sparsity} before. However, the planning and exploration techniques in these algorithms are difficult to scale. A key insight that \citet{lamb2022guaranteed} uncovered is the usefulness of multi-step action prediction for learning exogenous-invariant representation. However this work was limited to settings where the endogenous dynamic is tabular and contains small amount of latent states. Further, they did not use the learned representations to solve a downstream task.  

\begin{figure*}[t!]
    \centering
    \includegraphics[trim={0cm 0cm 0cm 1.5cm}, clip, width=0.405\textwidth]{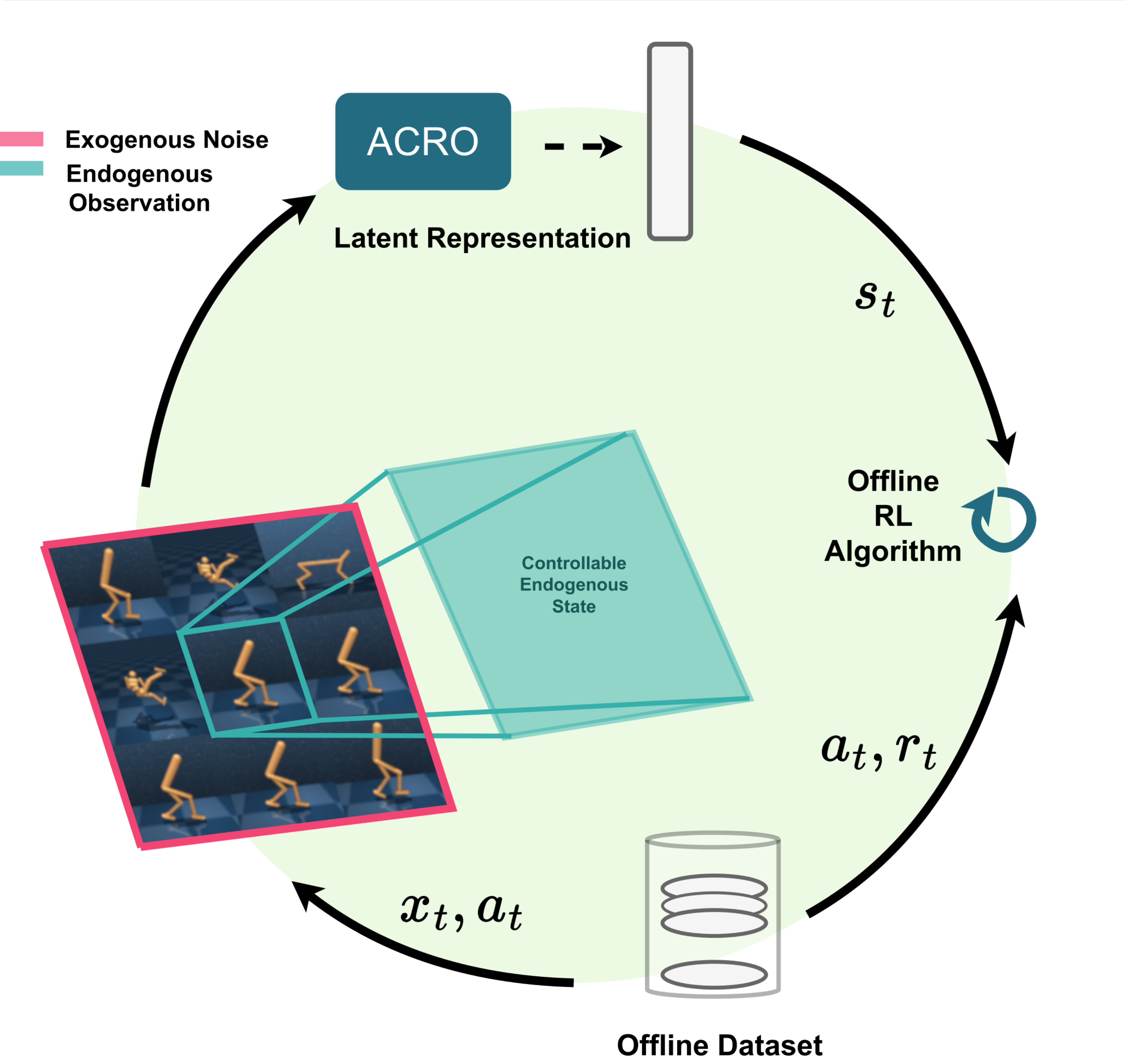}
    \includegraphics[trim={0cm 0cm 0cm 0cm}, clip, width=0.468\linewidth]{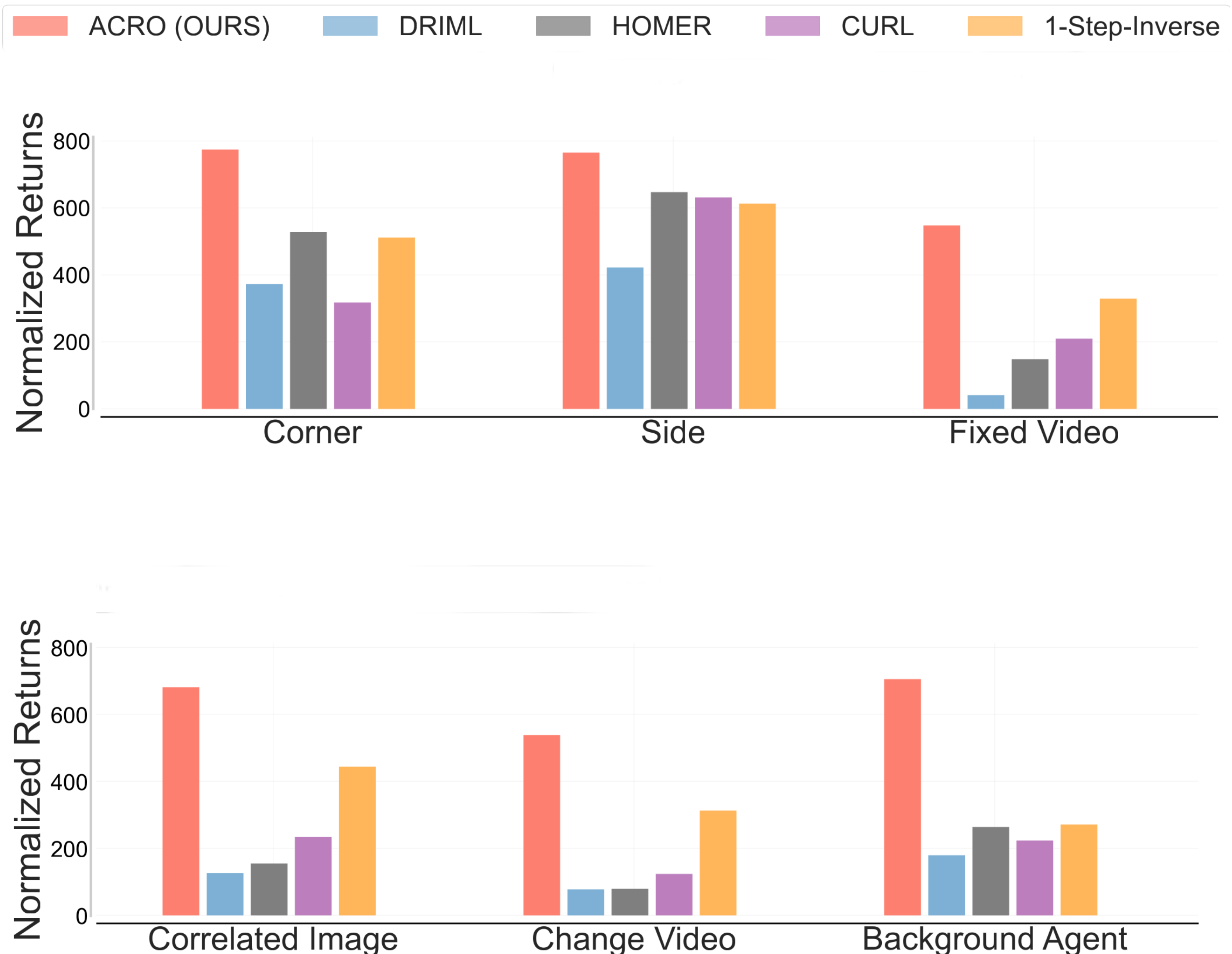} \\
    \caption{\textbf{Left: Representation Learning for Visual Offline RL in Presence of Exogenous Information}. We propose $\methodname$, that recovers the agent-centric latent representations from visual data which includes uncontrollable irrelevant information, such as observations of other agents acting in the same environment. \textbf{Right: Normalized Results Summary}. $\methodname$ learns to ignore the observations of task irrelevant agents, while baselines tend to capture such exogenous information. We use different offline datasets with varying levels of exogenous information (\pref{sec:exps_details}) and find that baseline methods consistently under-perform w.r.t. $\methodname$, as is supported by our theoretical analysis. Experimental results are normalized (averaged) across domains and different types of datasets (expert, medium-expert and medium). }
    \label{fig:motivation}
\end{figure*}

We propose to learn \textit{Agent-Centric Representations for Offline-RL (ACRO)} using multi-step inverse models, which predict actions given current and future observations, as in~\pref{fig:ACRO_Illustration}.  \methodname avoids the problem of learning distractors because they are not predictive of the agent's actions.  This property even holds for temporally-correlated exogenous information. At the same time, we show that multi-step inverse models capture all the information that is sufficient for controlling the agent while being entirely reward-free, which we refer to as the agent-centric representation. Our first contribution is to show that \methodname outperforms all current baselines on datasets from policies of varying quality and stochasticity. \pref{fig:motivation} gives an illustration of \methodname along with a summary of experimental findings.

A second core contribution of this work is to develop and release several new offline-RL benchmarks designed to have especially challenging exogenous information. In particular, we focus on \textit{diverse temporally-correlated} exogenous information with datasets where (1) every episode has a different video playing in the background, (2) the same STL-10 image is placed to the side or corner of the observation throughout the episode, and (3) the observation consists of the views of nine independent agents but the actions only control one of them (see Fig.~\ref{fig:motivation}). Task (3) is  challenging since the agent that is controllable must be learned from data. 

Finally, we also introduce a new theoretical analysis (\pref{sec:theory}) that explores the connection between exogenous noise in the learned representation and the success of Offline-RL. In particular, we show that Bellman completeness is achieved from the agent-centric representation of \methodname while representations which include exogenous noise may not verify it. Bellman completeness has been previously shown to be a sufficient condition for the convergence of offline RL methods based on Bellman error minimization \citep{munos2003error,munos2008finite}. This highlights the challenges of Offline RL with exogenous information at the observation level.  

\section{ACRO: Agent-Centric Representations for Offline-RL}
\vspace{-1mm}

\subsection{Preliminaries} 
\vspace{-2mm}
\label{sec:preliminaries}
We consider a Markov Decision Process (MDP) setting for modeling systems with both relevant and irrelevant components (also referred as exogenous block MDP in \citet{efroni2021provably}). This MDP consists of a set of observations, $\mathcal{X}$; a set of latent states, $\mathcal{Z}$; a set of actions, $\mathcal{A}$; a transition distribution, $T(\sff' \mid \sff, a)$; an emission distribution $q(x \mid \sff)$; a reward function $R: \Xcal \times \Acal \rightarrow \mathbb{R}$; and a start state distribution $\mu_0(z)$. We also assume that the support of the emission distributions of any two latent states are disjoint.  The latent state is decoupled into two parts $\sff =(\sd,\sx)$ where $\sd \in \mathcal{S}$ is the agent-centric state and $\sx\in \mathcal{E}$ is the exogenous state. For $\sff,\sff'\in \Sf,a\in \Acal$ the transition function is decoupled as
    $
       T(\sff' \mid \sff, a) = T(\sd' \mid \sd, a) T_{\sx}(\sx' \mid \sx),
     $

and the reward only depends on $(s,a)$. These definitions imply that there exist mappings $\phi_\star:\Xcal\rightarrow \mathcal{S}$ and $\phi_{\star,e} :\Xcal\rightarrow \mathcal{E} $ from observations to the corresponding agent-centric and exogenous and uncontrollable latent states. As in~\citet{lamb2022guaranteed}, we assume that the agent-centric dynamics is deterministic. The agent interacts with the environment, generating a latent state, observation and action sequence, $(\sff_1, x_1, a_1, \sff_2, x_2, a_2,\cdots,)$ where $\sff_1 \sim \mu(\cdot)$ and $x_t \sim q(\cdot \mid \sff_t)$. The agent does not observe the latent states $\rbr{\sff_1,\sff_2, \cdots}$, instead receiving only the observations $\rbr{x_1,x_2,\cdots}$.  The agent chooses actions using a policy distribution $\pi(a \mid x)$. A policy is an \textit{exo-free policy} if it is not a function of the exogenous noise. Formally, for any $x_1$ and $x_2$, if $\phi_\star(x_1) = \phi_\star(x_2)$, then $\pi(\cdot \mid x_1) =  \pi(\cdot \mid x_2)$.

We consider learning representations from an offline dataset $\mathcal{D} = (\mathcal{X}, \mathcal{A})$ consisting of sequences of N observations $\mathcal{X} = (x_1, x_2, x_3, ..., x_N)$ and the corresponding actions $\mathcal{A} = (a_1, a_2, a_3, ..., a_N)$. We are in the rich-observation setting, \textit{i.e.}, observation $x_t \in \mathbb{R}^m$ is sufficient to decode $z_t$. Our focus is on pre-training an encoder $\phi: \mathbb{R}^m \rightarrow \mathbb{R}^d$ on $\mathcal{D}$ such that the frozen representation $s_t = \phi(x_t)$ is suitable for offline policy optimization.  In our setting, we assume that the reward function is free of exogenous noise, and only depends on the endogenous part of the observation space.

\subsection{Benefits of Exogenous Invariant Representation in Offline RL}
\vspace{-1mm}
\label{sec:theory}
Due to its importance to practical applications, the offline RL setting has been extensively studied by the theoretical community. The majority of provable value-based offline RL algorithms follow a  Bellman error minimization approach~\citep{munos2003error,munos2008finite,antos2008learning}, in line with the techniques used in practice. 
The common representational assumptions needed to derive these results are:
(A1) the function class contains the optimal Q function (realizability), (A2) the data distribution is sufficiently diverse (concentrability), and (A3) Bellman completeness~\citep{munos2008finite}. This last condition states that the function class can properly represent the Bellman backup of any function it contains.

\begin{definition}[Bellman Completeness] We say that a function class $\Fcal$ is Bellman complete if it is closed under the Bellman operator. That is, for any $f\in \Fcal$ it holds that $\Tcal f\in \Fcal$, where $(\Tcal f)(x,a)\equiv R(x,a) + \EE_{x'\sim T(x'\mid x,a)}[ \max_{a'} f(x',a')]$ for all $(x,a)\in \Xcal\times \Acal$. 
\end{definition}

\citet{chen2019information} conjectured that (A1) and (A2) alone are not sufficient for sample efficient offline RL, and, recently,~\citet{foster2021offline} established a lower bound proving this claim. Thus, the representational requirements needed for offline RL are more intricate than in supervised learning. 

With these observations in mind, we highlight a key advantage of the agent-centric representation $\phi_\star$ relatively to other representations in the offline RL setting. Namely, we show one can construct a Bellman complete function class on top of $\phi_\star$, while some representations that include exogenous information provably violate Bellman completeness. To formalize these claims, we denote by $\mathcal{Q}_{\Scal} = \{ (s,a) \mapsto [0,1]: (s,a)\in \mathcal{S} \times \mathcal{A} \}$ the set of Q-functions defined over $\mathcal{S}$, and for a given representation $\phi$, we let $\Fcal(\phi) = \{ (s,a) \mapsto  Q(\phi(s),a) : Q \in \mathcal{Q}_{\Scal}, (s,a)\in \mathcal{S} \times \mathcal{A} \}$ denote the set of Q-functions defined on top of $\phi$. The following proposition states that the Agent-Centric representation leads to a Bellman complete function class (all proofs in~\pref{app:prop3}/~\pref{app:prop4}).

\begin{proposition}[\methodname Representation is Bellman Complete] \label{prop:controllable_rep_advantage}
$\Fcal(\phi_\star) $ is Bellman complete.
\end{proposition}
Next, we show that there exists a representation strictly more expressive than \methodname (\textit{i.e.}, one that includes exogenous information and all the agent-centric information) which, surprisingly, violates the Bellman completeness property.
\begin{proposition}[Exogenous Information May Violate Bellman Completeness]  \label{prop:exo_rep_disadvantage}
There exists $\phi$ which is a refinement\footnote{Let $\Xcal$ be a finite set of elements. Given a partition $P$ of $\Xcal$ let its induced equivalence relation be denoted by $\sim_P$. A partition $P_1$ is finer than $P_2$ if for any $x_1,x_2\in \Xcal$ such that $x_1 \sim_{P_1} x_2$ it also holds that $x_1 \sim_{P_2} x_2$.
} of $\phi_\star$ such that $\Fcal(\phi) $ is not Bellman complete.
\end{proposition}

This proposition implies that exogenous information being included in the representation may break the Bellman completeness assumption, which is a requirement for establishing the convergence of offline RL algorithms based on Bellman error minimization. From this perspective, additional information in the representation may deteriorate the performance of offline RL. Conversely, a coarser representation may trivially violate the realizability assumption A1: such a representation may merge states on which the optimal Q-function differs, preventing it from being realized. 

Together, these observations motivate the experimental pipeline
used this work: learn the agent-centric representation by optimizing~\pref{eqn:acro}, then perform offline RL on top of it. In doing so, we obtain a representation that is sufficient for optimal performance, and yet filters the exogenous information which can (i) be impossible to exactly model, and (ii) hurt the offline RL performance.

\subsection{Proposed Method: ACRO}
\vspace{-1mm}
\label{sec:acro_method}
\begin{figure}
    \centering
    \includegraphics[trim={0cm 0cm 0cm 0cm}, clip, width=0.5\linewidth]{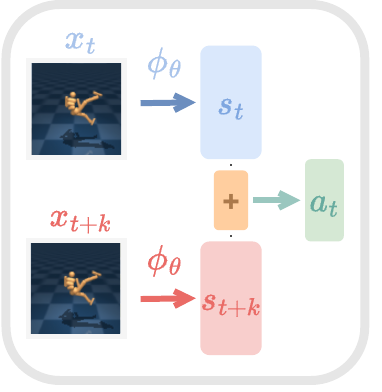}
    \caption{\textbf{$\methodname$}. is a multi-step inverse model that predicts the first action conditioned on the current state and the future state. \textbf{+} denotes concatenation. }
    \label{fig:ACRO_Illustration}
\end{figure}

\textbf{Learning Representations}: To learn representations that discard exogenous information, we leverage prior works from the theoretical RL community and train a multi-step inverse action prediction model, which captures long-range dependencies thanks to its conditioning on distant future observations. This leads to the \methodname objective, namely to predict the action conditioning on $\phi(x_t)$ and $\phi(x_{t+k})$. Note that even though we are conditioning on the future observation, we only predict the first action instead of the sequence of actions up to the k-th timestep, as the former is easier to learn ($k$ is sampled up to a maximum span of $K$). 

Our proposed method, which we call \textit{Agent-Centric Representations for Offline-RL} ($\methodname$), optimizes the following objective based on a multi-step inverse model:
\begin{align}
\label{eq:acro_objective}
\phi_\star \in \argmax_{\phi \in \Phi} \E_{\substack{t \sim U(0,N), \\ k \sim U(0,K)}}  \text{log}\left(\PP(a_t \mid \phi(x_t), \phi(x_{t+k}))\right) 
\end{align}
This approach (Figure~\ref{fig:ACRO_Illustration}) is motivated by two desiderata: (i) ignoring exogenous information and (ii) capturing the latent state that is necessary for control. The invariance lemma \citep{efroni2021provably} (see~\pref{app:factorizationlemma} for proof) states that optimal action predictor models can be obtained without dependence on exogenous noise, when the data-collection policy is assumed not to depend on it either.

At the same time, prior work has shown that single-step inverse models of action prediction can fail to capture the full agent-centric latent state \citep{efroni2021provably,lamb2022guaranteed,hutter2022uniqueness}.  One type of counter-example for single-step inverse models stems from a failure to capture long-range dependencies.  For example, in an empty gridworld, a pair of positions that are two or more spaces apart can be mapped to the same representation without increasing the loss of a one-step inverse model.  Another simple counter-example involves a problem where the last action the agent took is recorded in the observation, in which case the encoder can simply retrieve that action directly while ignoring all other information (although recording all recent actions in the observation is an issue for multi-step inverse models). The use of multi-step inverse models resolves both of these counter-examples and is able to learn the full agent-centric state \citep{efroni2021provably}. Detailed theoretical analysis on \methodname provided in appendix~\ref{app:acro_theory}, with specific counterexamples to one-step models in ~\ref{sec:counterexample}.

We emphasize here that even though inverse models of action prediction have appeared in past literature (as discussed in related works), they are often proposed for the purposes of exploration and reward bonus. In contrast, we propose to learn the multi-step inverse model to explicitly uncover a representation that contains only the agent-centric, endogenous part of the state. Recently, \citet{lamb2022guaranteed} proposed a multi-step inverse model where the learnt representation $\phi(\cdot)$ is regularized, so that $\phi(\cdot)$ discards irrelevant details from observations $x$. This was accomplished by using vector-quantization on the encoder's output, forcing discrete latent states to be learnt for constructing a tabular MDP for latent recovery. In contrast, \methodname is not limited to tabular settings and learns a continuous endogenous latent state without a bottleneck.  The learnt pre-trained representation $\phi(\cdot)$ is used for policy optimization in offline RL. More details of our algorithm are discussed in~\pref{app:exp_setup}.

\paragraph{Offline RL:} Given the learnt representation $\phi$, we can then use any existing offline RL algorithm. The performance on the downstream task depends on the robustness of the learnt representation $\phi$. For our experiments, we build off from the open source code base accompanying the v-d4rl benchmark~\citep{vd4rl}. We implement the pre-trained representation objectives in a model-free setting, where we use \textbf{TD3 + BC} as the baseline offline RL algorithm \citep{td3bc}. The policy improvement objective for the baseline RL algorithm is given by: $    L_{\theta}(\mathcal{D}) = - \mathbb{E}_{x_t \sim \mathcal{D}_{\phi}} \Big[   Q_{\psi} (s_t, \tilde{a}_t)\Big] $ where $s_t = \phi(\text{aug}(x_t))$ is the encoded augmented visual observation, $\tilde{a}_t = \pi_{\theta}(s_t) + \epsilon$ (action with clipped noise to smooth targets, $\epsilon \sim \text{clip}( \mathcal{N}(0, \sigma^2), -c, c)  )$.  The critic $Q(s, \pi(s))$ is evaluated by a TD loss, and we use re-parameterized gradients through the critic for policy improvement step. The overall loss for policy improvement using the encoded $\mathcal{D}_\phi$ is: 
\begin{equation}
    \mathcal{L}_{\theta}(\mathcal{D}) = - \mathbb{E}_{s_t, a_t \sim \mathcal{D}_\phi} \Big[ \lambda Q_{\psi}(s_t, \rva_t) - ( \pi_{\theta}(s_t) - \rva_t )^{2}  \Big] 
\end{equation}
For pixel based visual observations, recent work \cite{vd4rl} used TD3 algoroithm along with \textbf{DrQ + BC}, where it additionally applies the data augmentations on pixel based inputs. DrQ passes the gradients of the critic to learn the encoder, and there are no separate or explicit representation losses other than the critic estimation, for training the encoder in DrQ. We use the same offline experiment pipeline from \cite{lu2022challenges}, where representations are pre-trained with \methodname.

\vspace{-0.8em}
\subsection{Does \methodname also remove task relevant information?}
We produced a small analysis to demonstrate that the agent-centric representation captures information about objects that the agent can affect.  Let us consider two variants of a gridworld environment in which actions move the agent in the four directions (left, right, up, and down), and the agent is able to move a block without the block itself having any effect on the agent. In a push variant, the agent moves the block when it moves towards it (moving off the edge of the gridworld wraps onto the other side, to prevent the block from getting trapped in the corners).  In a pull variant, the agent moves the block along with itself unless it moves to the outer edge of the grid, which causes the agent to drop the block.  We trained \methodname on 500000 samples on each of these environments, and found that the learnt representation captures the state of both the agent and the block, while discarding exogenous noise.  Detailed experimental results extensive discussion  included in Appendix~\ref{app:blockworld}.  

This demonstrates that the agent-centric representation is more extensive than might be imagined at first glance. Anything that influences the actions taken by the policy needs to be captured by the representation in order to predict the first action it took from a pair of representations. As an additional illustration, consider the example of a robot hand grabbing a block, and a pair of states $(s_t, s_{t'})$, where $s_t$ corresponds to the state before the object was grasped and $s_{t'}$ after. The agent-centric representation needs to include information that pertains to the block's position and orientation; otherwise, it would be impossible to predict the actions governing the robot's motion (in the direction of the object) or how its joints are adjusting to grab the object. Note that this would not be required for a simple one-step inverse model, as the robot hand's joint positions in successive states are sufficient to infer what action was taken.

\section{Related Work}
\vspace{-1mm}
\label{sec:related}

In~\pref{tab:related_works}, we list prior works and whether they verify various properties, in particular invariance to exogenous information. An extended discussion on related works is provided in~\pref{app:related_work}. 

\begin{table}
\caption{\textbf{Overview of Properties}. of prior works on representation learning in RL, in particular their robustness to exogenous information. The comparison to \methodname aims to be as generous as possible to the baselines. \xmark\, is used to indicate a known counterexample for a given property. We compare the properties (i) Time-Independent Exogenous Invariant (ii) Reward-Free (iii) Exogenous Invariant (iv) Non-Expert Policy (v) Full Agent-Centric Representation.}

\label{tab:related_works}

\centering
\resizebox{\linewidth}{!}{%

\begin{tabular}{l|c c c c c c c c c}

Algorithms & \multicolumn{1}{l}{\begin{tabular}[c]{@{}l@{}}$\textsc{TD3}$ \\ (DrQ)\end{tabular}} & \multicolumn{1}{l}{\begin{tabular}[c]{@{}l@{}}\textsc{CURL}\end{tabular}} & \multicolumn{1}{l}{\begin{tabular}[c]{@{}l@{}}DRIML\end{tabular}} & \multicolumn{1}{l}{\begin{tabular}[c]{@{}l@{}}DBC\end{tabular}} & \multicolumn{1}{l}{\begin{tabular}[c]{@{}l@{}}AE\end{tabular}} & \multicolumn{1}{l}{\begin{tabular}[c]{@{}l@{}}1-Step\\Inverse\end{tabular}} & \multicolumn{1}{l}{\begin{tabular}[c]{@{}l@{}}Behavior\\Cloning\end{tabular}} &
\multicolumn{1}{l}{\begin{tabular}[c]{@{}l@{}}BYOL \\ Explore\end{tabular}} &
\multicolumn{1}{l}{\begin{tabular}[c]{@{}l@{}}$\textbf{\methodname}$\\(Ours)\end{tabular}} \\ 
\shline
\makecell{Time-Ind.\\Exo. Inv.} & \bluecheck & \xmark & \bluecheck & \bluecheck &
\xmark & \bluecheck & \bluecheck & \bluecheck & \bluecheck \\
\makecell{Reward\\Free} & \xmark & \bluecheck & \bluecheck & \xmark & \bluecheck & \bluecheck & \bluecheck & \bluecheck & \bluecheck \\
\makecell{Exogenous\\Invariant} & \xmark & \xmark & \xmark & \bluecheck & \xmark & \bluecheck & \bluecheck & ? & \bluecheck \\
\makecell{Non-Expert\\Policy} & \bluecheck & \bluecheck & \bluecheck & \bluecheck & \bluecheck & \bluecheck & \xmark & \bluecheck & \bluecheck \\
\makecell{Full\\Rep.} & \bluecheck & \xmark & \bluecheck & \xmark & \bluecheck & \xmark & \bluecheck  & \xmark & \bluecheck
\end{tabular}
}

\end{table}

\textbf{Inverse Dynamics Models}. One-Step Inverse Models predict the action taken conditioned on the previous and resulting observations. This is invariant to exogenous noise but fails to capture the agent-centric latent state~\citep{efroni2021provably}, as previously discussed in ~\pref{sec:acro_method}. This can result from inability to capture long-range dependencies or could result from trivial prediction of actions using a dashboard displaying the last action taken, such as the brakelight which turns on after the break is applied on a car~\citep{de2019causal}. Behavior Cloning predicts actions given current state and may also condition on future returns.  This is invariant to exogenous noise but can struggle with non-expert policies and generally fails to learn agent-centric latent state. Inverse models predicting sequences of actions, like \textsc{GLAMOR}~\citep{paster2020planning} considers an online setting where they learn an action sequence as a sequential multi-step inverse model and rollout via random shooting and re-scoring, using both the inverse-model accuracies and an action-prior distribution. Additional discussion and experimental results comparing \methodname with action sequence prediction, are provided in appendix \ref{app:acro_sequence_action_prediction}. 

\textbf{Contrastive Methods}. \textsc{CURL} (Augmentation Contrastive, ~\cite{laskin2020curl}) learns a representation which is invariant to a class of data augmentations while being different across random example pairs. Depending on what augmentations and datasets are used, the learnt representations would generally learn exogenous noise and also fail to capture agent-centric latent states (which could be removed by some augmentations). \textsc{HOMER} and \textsc{DRIML} (Time Contrastive, \cite{misra2020kinematic}, \cite{mazoure2020deep}) learns representations which can discriminate between adjacent observations in a rollout and pairs of random observations. This has been proven to not be invariant to exogenous information and neither can capture the agent-centric latent state~\citep{efroni2021provably}. 

\textbf{Predictive Models}. Autoencoders learn to reconstruct an observation through a representation bottleneck. Generative modeling approaches usually capture all information in the input space which includes both exogenous noise and the agent-centric latent state~\citep{hafner2019dream}.  \citet{wang2022denoised,wang2022causal} showed that a generative model of transition in the observation space can decompose the space into agent-centric state and exogenous information. While this does, in principle, eventually achieve an Agent-Centric representation, it comes at the cost of learning the exogenous representation and its dynamics before discarding the information. \textsc{BYOL-Explore} \citep{guo2022byolexplore} achieved impressive empirical results in online exploration by predicting future representations based on past representations and actions. While this approach can ignore exogenous information, there is no guarantee that it will do so, nor that it will learn the full agent-centric state.  

\textbf{RL with Exogenous Information}. Several prior works study the RL with exogenous information problem. In~\citet{dietterich2018discovering,exoRL,efroni2022sparsity}, the authors consider specific representational assumptions on the underlying model, such as linear dynamics or factorized representation of the exogenous information in observations. Our work focuses on the rich observation setting where the representation itself should be learned. \citet{efroni2021provably} proposes a deterministic path planning algorithm for being invariant to exogenous noise. Unlike their approach which requires interaction with the environment using a tabular-MDP, \methodname is a purely offline algorithm. Lastly, the work of \citet{lamb2022guaranteed}   suggests an endogenous latent state recovery algorithm through the use of a discretization bottleneck. Their approach is designed to work for MDPs with tabular and small endogenous state space. In contrast, \methodname recovers a continuous counterpart of the endogenous latent state space directly, without the need to construct a tabular-MDP. Hence, it is applicable for larger scale problems. Furthermore, we focus on reward optimization, not only on latent state discovery.

\section{Experiments: Offline RL with Exogenous Information}
\vspace{-1mm}
\label{sec:exps_details}

\begin{table*}[t!]
\caption{ \textbf{\easyexo}. Comparison of different representation methods on the standard v-d4rl benchmark, without additional exogenous information. \methodname consistently outperforms baseline methods in visual offline data. Performance plots in Appendix~\pref{fig:vd4rl_main_pre_train}. 10 seeds and std. dev. reported.}\label{tab:easy_exo_results}
\resizebox{\linewidth}{!}{%
\tablestyle{8pt}{1.1}
\begin{tabular}{ll|c c c c c c c}
\label{tab:easy_exo_without_distractors}
\textsc{Environment} & \textsc{Dataset} &
  \textsc{ACRO} &
  \textsc{DRIML} &
  \textsc{HOMER} &
  \textsc{DrQv2} &
  \textsc{CURL} &
  \textsc{1-Step Inverse} \\ 
\shline
\multicolumn{1}{l|}{\multirow{4}{*}{\textsc{Cheetah-Run}}} &
  Expert &
  \highlight{451.0 $\pm$ 3.9} &
  330.2 $\pm$ 2.9 &
  227.8 $\pm$ 1.6 &
  256.9 $\pm$ 2.2 &
  213.0 $\pm$ 0.6 &
  239.9 $\pm$ 0.4 \\ 
\multicolumn{1}{l|}{} & Medium-Expert & \highlight{466.0 $\pm$ 3.2}  & 399.2 $\pm$ 2.5 & 390.7 $\pm$ 1.3 & 388.1 $\pm$ 3.5 & 328.4 $\pm$ 2.1  & 299.3 $\pm$ 0.6\\ 
\multicolumn{1}{l|}{} & Medium        & \highlight{528.7 + 0.8}  & 508.5 $\pm$ 0.7 & 518.1 $\pm$ 0.4 & 488.3 $\pm$ 0.5 & 377.0 $\pm$ 0.8  & 400.3 $\pm$ 0.4 \\ 
\multicolumn{1}{l|}{} & Medium-Replay & \highlight{416.9 $\pm$ 0.9}  & 233.3 $\pm$ 1.2 & 333.2 $\pm$ 1.2 & 381.5 $\pm$ 1.6 & 279.4 $\pm$ 1.8  & 272.3 $\pm$ .6  \\ \hline
\multicolumn{1}{l|}{\multirow{4}{*}{\textsc{Walker-Walk}}} &
  Expert &
  \highlight{924.5 $\pm$ 2.2} &
  485.10 $\pm$ 4.9 &
  670.55 $\pm$ 4.1 &
  888.6 $\pm$ 6.0 &
  800.36 $\pm$ 2.5 &
  831.5 $\pm$ 3.4 \\ 
\multicolumn{1}{l|}{} & Medium-Expert & \highlight{914.6 $\pm$ 1.8} & 438.3 $\pm$ 3.3 & 774.5 $\pm$ 2.5 & 906.6 $\pm$ 0.9 & 724.6 $\pm$ 4.5  & 651.8 $\pm$ 4.0\\ 
\multicolumn{1}{l|}{} & Medium        & \highlight{486.7 $\pm$ 0.2}  & 469.4 $\pm$ 0.5 & 485.1 $\pm$ 0.7 & 425.6 $\pm$ 1.6 & 429.0 $\pm$ 2.0  & 389.4 $\pm$ 1.1 \\ 
\multicolumn{1}{l|}{} & Medium-Replay & 277.8 $\pm$ 0.5  & 204.3 $\pm$ 3.4 & \highlight{318.9 $\pm$ 4.0} & 308.5 $\pm$ 1.5 & 234.8  $\pm$ 2.4 & 146.7 $\pm$ 0.7 \\ \hline
\multicolumn{1}{l|}{\multirow{4}{*}{\textsc{Humanoid-Walk}}} &
  Expert &
  \highlight{79.9 $\pm$ 1.1} &
  17.5 $\pm$ 0.1 &
  21.6 $\pm$ 0.4 &
  34.1 $\pm$ 0.3 &
  28.5 $\pm$ 0.2 &
  25.4 $\pm$ 0.1 \\ 
\multicolumn{1}{l|}{} & Medium-Expert & \highlight{142.4 $\pm$ 1.2}  & 26.8 $\pm$ 0.2  & 31.8 $\pm$ 0.1  & 70.8 $\pm$ 0.5  & 63.2 $\pm$ 0.9   & 56.3 $\pm$ 0.5  \\ 
\multicolumn{1}{l|}{} & Medium        & \highlight{103.8 $\pm$ 1.8}  & 35.1 $\pm$ 0.3  & 53.8 $\pm$ 0.4  & 96.4 $\pm$ 0.9  & 40.6 $\pm$ 0.4    & 46.7 $\pm$ 0.1 \\ 
\multicolumn{1}{l|}{} & Medium-Replay & \highlight{197.8 $\pm$ 0.5}  & 92.6 $\pm$ 0.3  & 102.7 $\pm$ 0.6 & 121.0 $\pm$ 0.4 & 77.8 $\pm$ 0.8   & 100.7 $\pm$ 1.1  \\ \hline
\multirow{1}{*}{\textsc{Average}}
&  & \highlight{415.8}  & 270.0 & 327.4 & 363.9 & 299.7   & 288.4 
\\ 
\end{tabular}}
\end{table*}

This section provides extensive analysis of representation learning from visual offline data under rich exogenous information (\pref{fig:illustration_experiment_setups}). Our experiments aim to understand the effect of exogenous information and if \methodname can truly learn the agent-centric state and thus improve performance in visual offline RL. To this end, we evaluate $\methodname$ against several state of the art representation learning baselines across two axes of added exogenous information: \textit{Temporal Correlation} and \textit{Diversity}, hence characterizing the level of difficulty systematically. We find that under exogenous information in offline RL, the performance of several state of the art representation learning objectives can degrade dramatically.

Two particular challenges in the datasets we explore are the temporal correlation and diversity in the exogenous noise. \textit{Temporal Correlation:} Exogenous noise which lacks temporal correlation (time-independent noise) is relatively easy to filter out in the representation, especially in tasks where the agent-centric latent state has strong temporal correlation. \textit{Diversity:} Similarly for the other axis, if exogenous noise is more diverse, it has a greater impact on the complexity of the subsequently learned representation. For example, if there are only two possible distracting background images, in the worst case the cardinality of a discrete representation is only doubled. On the other hand if there are thousands of possible distracting background images, then the effect on the complexity of representation would be far greater. We primarily categorize our novel visual offline datasets into \textit{three categories} (\pref{fig:illustration_experiment_setups} in appendix provides observations under different exogenous distractors):

\begin{figure}[h]
\centering
\includegraphics[width=0.85\linewidth]{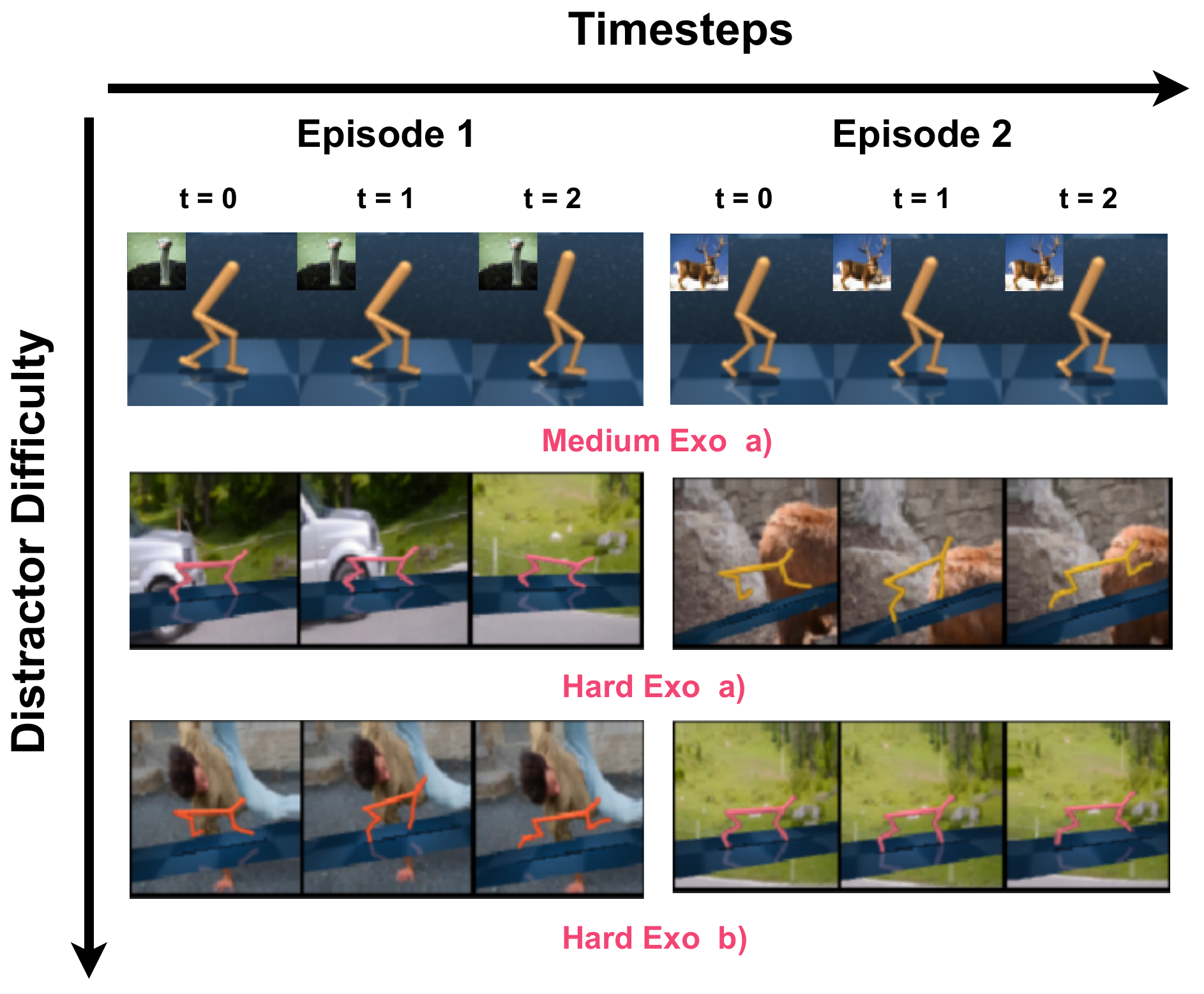}
\vspace{-2mm}
\caption{\textbf{Examples of Different Categories of Exogenous Information}. A simple set of distractor datasets are released at the following \href{https://drive.google.com/drive/folders/1HsksquQ6gKQUDj_1Qe7dc-R_h8m6A1H3?usp=drive_link}{GDrive link}.}
\label{fig:illustration_experiment_setups}
\end{figure}

\begin{itemize}[leftmargin=0.3cm]
    \item \textbf{\easyexo}. Exogenous noise with low-diversity and no time correlation. \textbf{a)} Visual offline datasets from v-d4rl benchmark \citep{lu2022challenges} without any background distractors; \textbf{b)} Distractor setting \citep{vd4rl} with a single fixed exogenous image in the background.
    \item \textbf{\mediumexo}. Exogenous noise with either low-diversity or simple time-correlation. \textbf{a)} Exogenous image placed in the corner of agent observations, changes per episode; \textbf{b)}  Exogenous image placed on the side of agent observations, changes per episode; \textbf{c)} A single fixed exogenous video playing in the background.
    \item \textbf{\hardexo}. Exogenous noise with both high-diversity and rich temporal correlation. \textbf{a)} Exogenous image in the background which changes per episode; \textbf{b) } Exogenous video in the background which changes per episode; \textbf{c)} Exogenous observations of nine agents placed in a grid, but the actions only control one of the agents (see~\pref{fig:motivation}).
\end{itemize}

\begin{figure*}[t!]
\includegraphics[width=1.0\textwidth]{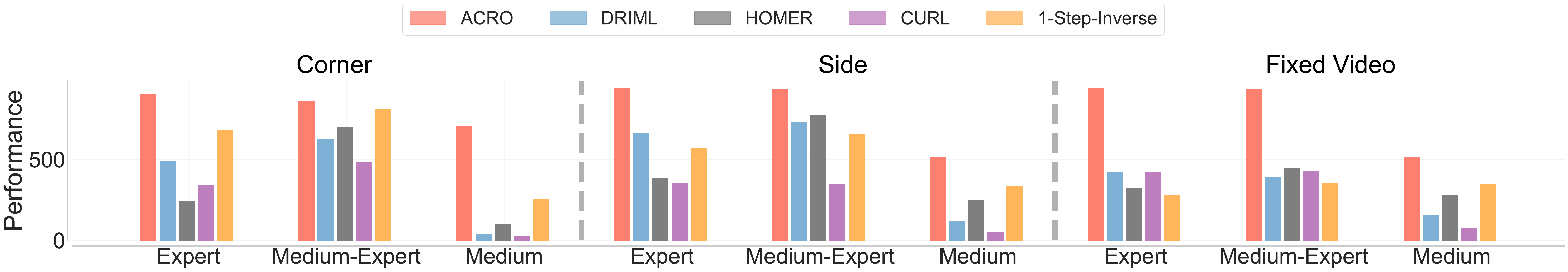}
\caption{\textbf{\mediumexo Results}. Performance comparison of $\methodname$ with several other baselines, with varying levels of exogenous information settings, either from STL10 dataset \citep{stl10} or fixed video distractors in background during offline data collection. Normalized (averaged) performance plots across different control domains (Humanoid, HalfCheetah and Walker) as we vary the type of offline data collecting policies.}
\vspace{-2mm}
\label{fig:medium_exo_results}
\end{figure*}

\textbf{Experiment Setup}. We provide details of each \exodatasets in~\pref{app:exp_setup}, and release a simple set of distractor datasets \href{https://drive.google.com/drive/folders/1HsksquQ6gKQUDj_1Qe7dc-R_h8m6A1H3?usp=drive_link}{here}. Following \citet{fu2020d4rl, lu2022challenges}, we release these datasets for future use by the RL community. All experiments involve pre-training the representation, and then freezing it for use in an offline RL algorithm. We use $\textsc{TD3 + BC}$ as the downstream RL algorithm, along with data augmentations~\citep{kostrikov2020image}. Experiment setup and implementation details are discussed in ~\pref{app:exp_setup}. Additional experimental results are also provided in appendix \ref{app:additional_experiments}.

\begin{figure}[h]
\includegraphics[width=0.9\linewidth]{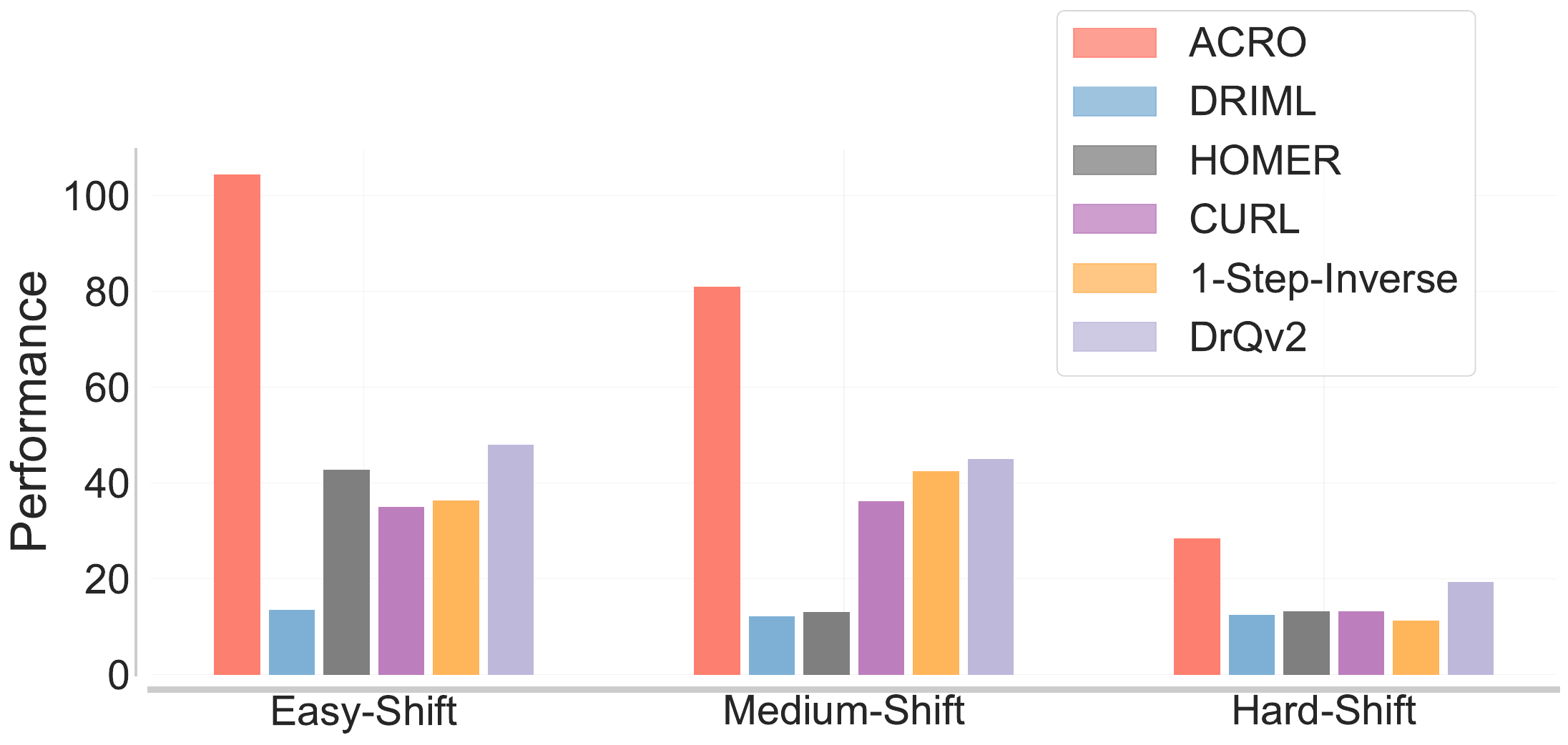}
\caption{\textbf{Normalized results}. across two domains from the v-d4rl distractor suite with varying levels (easy, medium and hard categories) of data shift severity \cite{vd4rl}. \vspace{-2.5em}}
\label{fig:easy_exo_barplot}
\end{figure}

\textbf{Baselines}. We compare \textit{five} baselines, which are standard for learning representations of visual data. The baselines we consider are: (i) two temporal contrastive learning methods, \textsc{DRIML}~\citep{mazoure2020deep} and \textsc{HOMER}~\citep{misra2020kinematic}; (ii) a data augmentation method, \textsc{DrQ}~\cite{kostrikov2020image}, and a spatial contrastive approach, \textsc{CURL}~\cite{laskin2020curl}; and (iii) inverse dynamics model learning, \textit{i.e.}, 1-step inverse action prediction~\citep{DBLP:conf/icml/PathakAED17}. We do not consider baselines such as \textsc{SPR}~\citep{schwarzer2020data} and \textsc{SGI}~\citep{SchwarzerSGI} which work well on the ALE Atari100K benchmark but not on control benchmarks~\citep{tomar2021learning}. We also include Atari results in Appendix~\ref{sec:atari-results} where representations are pre-trained using \methodname and used over a Decision Transformer~\citep{NEURIPS2021DT}.

\subsection{Easy-Exogenous Information Offline Datasets}
\vspace{-1mm}

\pref{tab:easy_exo_results} summarizes results from the v-d4rl benchmark with visual offline data \citep{vd4rl}. We label this as \easyexo since the dataset only contains a blank background without any additional exogenous noise being added. We find that \methodname learns a good agent-centric latent representation from pixel data with no apparent noise in observations, and can lead to effective performance improvements through pre-training representations. Extending results of \easyexo with static uncorrelated image background distractors from the v-d4rl benchmark, we see that the performance significantly decreases for all methods, while \methodname can strongly outperform all baselines, with the smallest drop in performance.~\pref{fig:easy_exo_barplot} shows normalized results across two different datasets and domains from v-d4rl. The distractors in this case belong to varying degree of shifts in the data distribution, according to~\citep{vd4rl}.

\subsection{Medium-Exogenous Information Offline Datasets}
\vspace{-1mm}
\pref{fig:medium_exo_results} shows normalized results across three domains (cheetah-run, walker-walk, humanoid-walk) for the \mediumexo setting. Among these, the fixed background video is the hardest task. Most methods underperform on data collected from a medium policy, compared to medium-expert and expert policies. However, \methodname consistently outperforms all methods across datasets and distractor settings. Note the high variability in performance of baselines when changing the type of exogenous information (from corner, to side, to fixed video), while in contrast, \methodname performs similarly for all three settings. This suggests that baseline methods do not learn exogenous-free robust representations, while \methodname remains impervious to it.

\subsection{Hard-Exogenous Information Offline Datasets}
\vspace{-1mm}

\begin{figure}[h!]
    \centering
    {\includegraphics[width=0.99\linewidth]{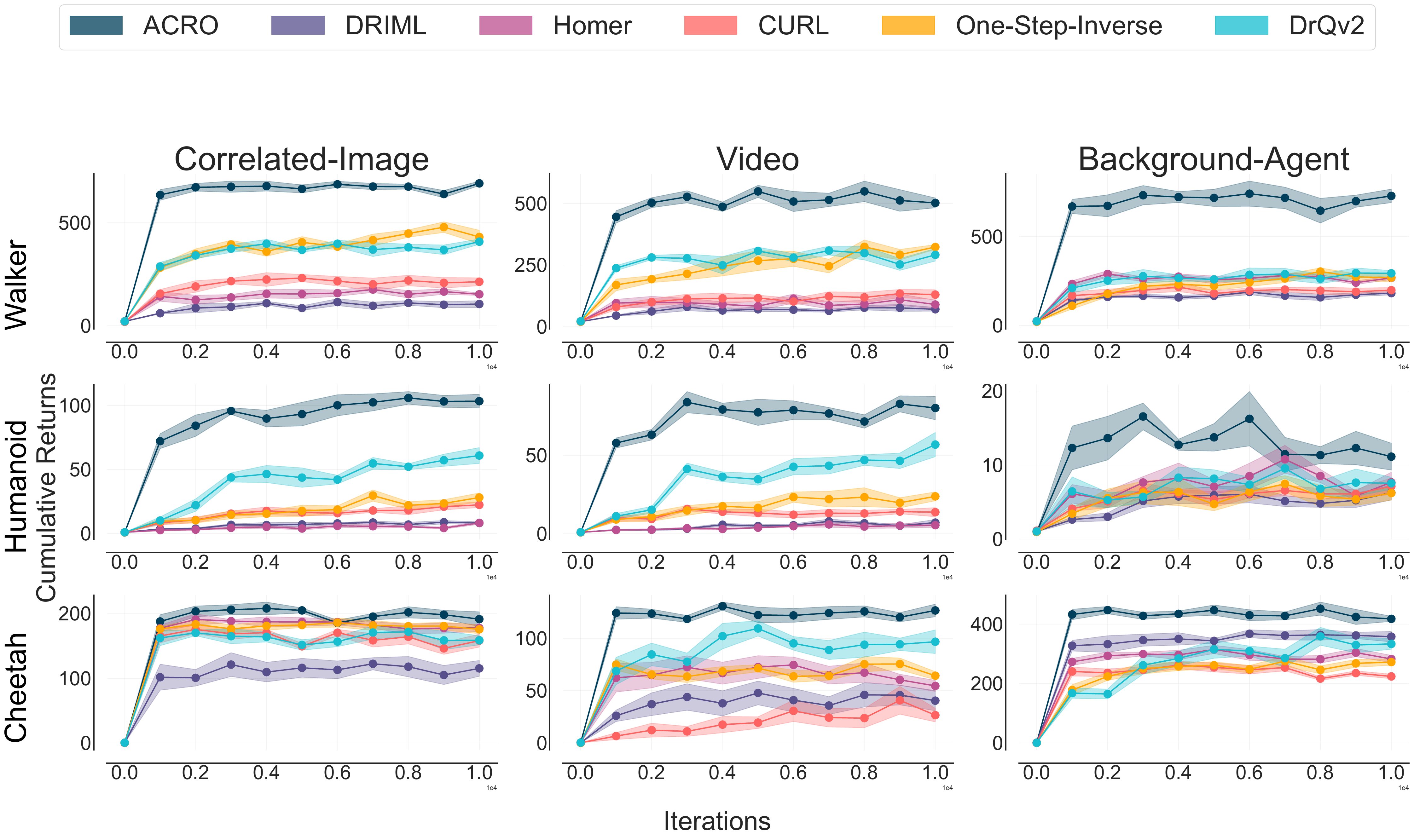}}
    \vspace{-2mm}
    \caption{\textbf{\hardexo Results}. Normalized performance across three datasets: medium, expert, and medium-expert. \textbf{First Column}. Time correlated exogenous distractor in background; \textbf{Second Column}. Video distractors that changes per episode in background; \textbf{Third Column}. Multiple background agent observations as distractors are placed in a grid of agent observation space. Normalized performance plots across different data collecting policies (expert, medium-expert and medium) for three different domains (Cheetah, Humanoid and Walker). \vspace{-1.2em}}
    \label{fig:hard_exo_results_multi_env}
\end{figure}

With correlated exogenous noise in the form of either images or video, we observe that baseline representation objectives can be remarkably broken.~\pref{fig:hard_exo_results_multi_env} shows normalized performance comparisons across different types of datasets (expert, medium-expert, medium) for three different types of \hardexo settings. Comparatively, \methodname can be more robust to the hard exogenous distractors, even though as the \hardexo types increase in difficulty, the maximum performance reached by all methods can degrade. Among the three \hardexo settings, changing video distractors in background during data collection seems to be the hardest, leading to performance drops for most methods. This suggests there is a strong correlation issue between the representation and the video pixels, which breaks when the episode changes, hence leading to the worst scores across the three settings. However, \methodname remains comparatively robust and outperforms all baselines across all the \hardexo settings.

\subsection{Method Ablations}
\vspace{-1mm}
We compare $\methodname$ with three of its variations, 1) when $k=1$, i.e. a standard one-step inverse model; 2) with $\rvx_{t+k}$ not provided as input, i.e. simply training the representation with a behavior cloning loss; and 3) when $m(k)$ the timestep embedding, is additionally provided as input. Ablations are shown over three different policies: random, medium-replay and expert in~\pref{tab:ablations}. Appendix \ref{app:data_coverage} includes results of \methodname on role of data coverage in offline datasets.

\begin{table}[h!]
\caption{\textbf{Ablations for different policies}. The highlighted cells indicate where each variant fails to match \methodname's performance, hence showing that each component of \methodname is essential for consistently good performance. 5 seeds and std. dev. reported.}\label{tab:ablations}
\vspace{-1mm}
\footnotesize
\tablestyle{4pt}{1.2}
\resizebox{\columnwidth}{!}{%
\begin{tabular}{l|c c c}
\multicolumn{1}{l|}{\textsc{Environment}} &
  \textsc{Random} &
  \textsc{Medium-Replay} &
  \textsc{Expert} \\ 
\shline
\textsc{ACRO} & 82.9 $\pm$ 5.5 & 228.8 $\pm$ 50.1 & 525.8 $\pm$ 89.0 \\
\textsc{K=1} & 94.7 $\pm$ 7.9 & 241.0 $\pm$ 9.9 & \redlight{187.5 $\pm$ 33.8} \\
\textsc{Only $x_t$} & \redlight{0.5 $\pm$ 0.1} & 229.4 $\pm$ 64.7 & 496.8 $\pm$ 100.2 \\
\textsc{With $k$} & 43.1 $\pm$ 49.5 & 251.8 $\pm$ 15.3 & \redlight{302.2 $\pm$ 29.1} \\
\end{tabular}
}
\end{table}
For both random and medium-replay policies, $k=1$ leads to similar results when $k$ is randomly chosen from 1 to 15. \methodname performs much better under an expert policy.  We conjecture that the benefits of larger $k$ can only be realized when the policy is of high enough quality to preserve information over long time horizons. Additionally, training a behavior cloning loss performs similarly to \methodname for the medium-replay and expert datasets.  However, when the actions come from a random policy, \methodname performs much better, while the behavior cloning ablation collapses completely.  This result is analyzed theoretically in~\pref{app:bceq}, which shows that \methodname is equivalent to behavior cloning under a deterministic and fixed expert policy, but should be much better otherwise.  Adding a $k$ embedding generally degrades performance, although the effect is inconsistent.  These results suggest that \methodname is a more well rounded and robust objective than other variants. In appendix 
\ref{app:rebuttal_generic_multi_action_prediction} we also include further ablations on how \methodname performs, when compared to predicting a sequence of actions.

\subsection{Visualizing Reconstructions from the Decoder}
\label{app:visualizations}

\begin{figure}[hp!]
\centering
\includegraphics[trim={0cm 0cm 0cm 0cm}, clip, width=\linewidth]{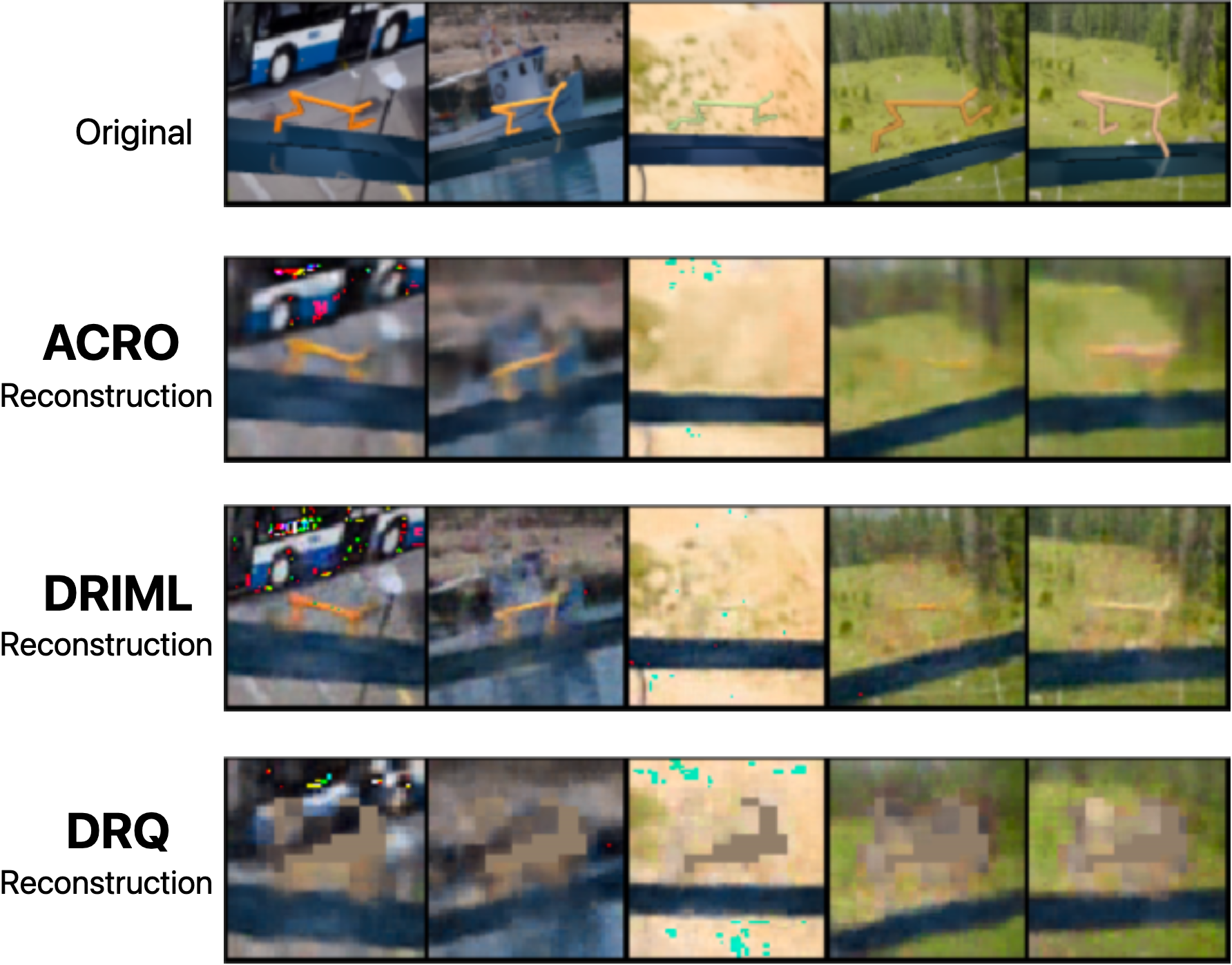}
\caption{\textbf{Reconstructions}. from a decoder with a static background image per episode: \textbf{Top-Bottom}: Original, \methodname, \textsc{DRIML}, \textsc{DrQ}.  Ideally we want to see that the background is blurry in the reconstruction, demonstrating removal of exogenous noise.  This is observed to some extent with both \textsc{ACRO} and \textsc{DrQ}, but not \textsc{DRIML}.  At the same time, we want to see that the agent is still visible in the reconstruction, which is mostly the case for \textsc{ACRO} and \textsc{DRIML}, but not \textsc{DrQ}}
\label{fig:recons}
\end{figure}

\textbf{Visualizing Reconstructions}. Having learnt a representation, we can train a decoder over it to minimize the reconstruction loss given the original observation. Such reconstructions would therefore measure how much information in the original observation is preserved in the representation, and thus act as a metric for evaluating the quality of representations. We compare such reconstructions in~\pref{fig:recons} for the cheetah domain where the exogenous noise comes from a video playing in the background. Notably, \methodname is able to remove most background information while keeping the relevant body pose information intact. On the other hand, \textsc{DRIML} performs contrastive comparisons between states in a given trajectory and is not able to remove exogenous information quite as well. \textsc{DrQ} is able to remove exogenous noise but is unable to learn the agent-centric state. Besides such qualitative differences in the learnt representations, we provide quantitative results showing how \methodname learns to remove exogenous information while retaining endogenous information in Appendix~\ref{app:quant-analysis}.

\vspace{-0.6em}
\section{Discussion}
\vspace{-1mm}
In this work, we introduced offline RL datasets with varying difficulties of exogenous information in the observations. Our results showcase that existing representation learning methods can significantly degrade performance for certain types of exogenous noise. We presented \methodname, a pre-training objective for offline RL based on a multi-step inverse prediction model, and showed it is far more robust to exogenous information, both theoretically and empirically. 

\textbf{Limitations and Future Work}. Since \methodname does not require reward information for learning representations, it is natural to wonder if data from multiple datasets (e.g., combining random and medium-replay) or different domains (e.g., a transfer learning setting) can be used to train a stronger representation than when using a single dataset. Additionally, under varying transition dynamics across tasks, a model-based counterpart of \methodname might be worth studying. For the domains and tasks considered in this work, it would be interesting to quantify how accurately \methodname recovers the underlying endogenous latent states, while using the latent structure for solving the task objective.

\section*{Acknowledgement}

The authors would like to thank Akshay Krishnamurthy, Dylan Foster, Romain Laroche, Michael
Janner, Anirudh Goyal and Doina Precup for valuable feedback on the draft. Hongyu Zang and Xin Li were partially supported by NSFC under Grant 62276024, 92270125, and U19B2020.

\bibliography{arxiv_icml}
\bibliographystyle{icml2023}

\newpage
\appendix

\onecolumn

\part*{Appendix}

\section{Benefits of Exogenous Invariant Representation in Offline RL}
\label{app:exo_theory}

\subsection{Proof of Proposition~\ref{prop:controllable_rep_advantage}.}
\label{app:prop3}
We need to show that for any $f\in \Fcal(\phi_\star)$ and $x,a$, it holds that
\begin{align}
    R(x,a) + \EE_{x'\sim T(\cdot\mid x,a)}[\max_{a'}f(x',a')] \label{eq: rep controllable adv 1}
\end{align}
is contained in $\Fcal(\phi_\star)$. Since the reward is a function of the agent controller representation only, and since $f\in \Fcal(\phi_\star)$,~\eqref{eq: rep controllable adv 1} can be written as:
\begin{align}
    &R(x,a) + \EE_{x'\sim T(\cdot\mid x,a)}[\max_{a'}f(x',a')] \nonumber\\
    &= R(\phi_\star(x),a) + \EE_{x'\sim T(\cdot\mid x,a)}[\max_{a'}f(\phi_\star(x'),a')] \nonumber  \\
    &= R(\phi_\star(x),a) + \EE_{x' \sim T(\cdot \mid \phi_\star(x),a)}[\max_{a'}f(\phi_\star(x'),a')]. \label{eq:proof_prop_endo}
\end{align}
The first relation holds since $f\in \Fcal(\phi_\star)$ and by the assumption on the reward function (that it is a function of the endogenous states). The second relation holds by
\begin{align*}
    &\EE_{x'\sim T(\cdot\mid x,a)}[\max_{a'}f(\phi_\star(x'),a')]\\
    &\stackrel{\mathrm{(a)}}{=} \sum_{s',e'} \sum_{x'\in \supp q(x' \mid s', e')} q(x' \mid \phi_\star(x'), \phi_{\star,e}(x'))T(s' \mid \phi_\star(x), a) T_{\sx}(e' \mid \phi_{\star,e}(x)) f(s',a')\\
    &\stackrel{\mathrm{(b)}}{=} \sum_{s'}T(s' \mid \phi_\star(x), a)  f(s',a')\sum_{e'}T_{\sx}(e' \mid \phi_{\star,e}(x))\\
    &\stackrel{\mathrm{(c)}}{=}\sum_{s'}T(s' \mid \phi_\star(x), a)  f(s',a'),
\end{align*}
where $\mathrm{(a)}$ holds by the Ex-BMDP transition model assumption,
\begin{align*}
    T(x'\mid x,a) =  q(x' \mid \phi_\star(x'), \phi_{\star,e}(x'))T(\phi_\star(x') \mid \phi_\star(x), a) T_{\sx}(\phi_{\star,e}(x') \mid \phi_{\star,e}(x)),
\end{align*}
$\mathrm{(b)}$ and $\mathrm{(c)}$ hold by marginalizing over $x'$ and $e'$. This establishes~\eqref{eq:proof_prop_endo} and the proposition: the function $R(\phi_\star(x),a) + \EE_{x' \sim T(\cdot \mid \phi_\star(x),a)}[\max_{a'}f(\phi_\star(x'),a')]$ is contained within $\Fcal(\phi_\star)$ since it only depends on $\phi_\star$.

\subsection{Proof of Proposition~\ref{prop:exo_rep_disadvantage}.}
\label{app:prop4}
Consider an Ex-BMDP with one action $a$ where the agent controller representation is trivial and has a single fixed state (the agent has no ability to affect the dynamics).  We will establish a counter-example by constructing a tabular-MDP.  Because tabular-MDP is a special case of a more general MDP with continuous states, this will also establish a counterexample for the more general non-tabular setting considered in the paper.  

Let the observations and dynamics be given has follows. The observation is a 2-dimensional vector $x= (x(1),x(2))$ where $x(1),x(2)\in \{ 0,1 \}$. The dynamics is deterministic and its time evoluation is given as follows:
\begin{align*}
    &x_{t+1}(1) = x_t(1) \oplus x_t(2)\\
    &x_{t+1}(2)=x_t(2),
\end{align*}
where $\oplus$ is the XOR operation. In this case, the transition model is given by $T(x'\mid x,a) = T(x'\mid x)$ and $\phi_\star= \{\ s_0\}$ where $s_0$ is a single state; since the observations are not controllable the controller representation maps all observations to a unique state. Further, assume that the reward function is $0$ for all observations.

Assume that $\phi(x) = (x_1)$, i.e., the representation ignores the second feature $x_2$. This representation is more refined than $\phi_\star$ since the latter maps all observations into the same state. Consider the tabular Q function class on top of this representation $\Qcal_{N=2}$, and consider $Q\in \Qcal_{N=2}$ given as follows
\begin{align*}
    &Q(x_1 = 1) = 1\\
    &Q(x_1 = 0) = 0.
\end{align*}

We now show that $\Tcal Q$ is not contained in $\Qcal_{N=2}.$ According to the construction of the transition model, it holds that
\begin{align*}
    &(T Q)(x_1=1,x_2=1) = 0\\
    &(TQ)(x_1=0,x_2=1) = 1\\
    &(T Q)(x_1=1,x_2=0) = 1\\
    &(TQ)(x_1=0,x_2=0) = 0.   
\end{align*}
This function cannot be represented by a function from $\Qcal_{N=2}$; we cannot represent $(T Q)$ since it is not a mapping of the form $x_1\rightarrow \mathbb{R}$ by the fact that, e.g.,  $$(T Q)(x_1=1,x_2=1) \neq (T Q)(x_1=1,x_2=0).$$ Meaning, it depends on the value of $x_2$.

\section{Theory on Learning Exogenous-Free Representations with ACRO}
\label{app:acro_theory}

\subsection{Multi-Step Inverse Model Invariance Proof}
\label{app:factorizationlemma}
For any $k>0$, consider $x,x'\in \mathcal{X}$ such that $x$ and $x'$ are separated by k steps.  Both proofs first use bayes theorem, then apply the factorized transition dynamics, and then eliminate terms shared in the numerator and denominator.  This proof is essentially the same as lemmas found in \cite{efroni2021provably,lamb2022guaranteed}, but is presented here for clarity.  

The multi-step inverse model is invariant to exogenous noise.  For any exo-free policy $\pi: \Xcal \rightarrow \Acal$, for all $a_t \in \Acal$, and $(x_t, x_{t+k}) \in \supp~\PP_\pi(X_t, X_{t+k})$:
\begin{equation}
    \PP_\pi(a_t \mid x_t, x_{t+k}) = \PP_\pi(a_t \mid \phi_\star(x_t), \phi_\star(x_{t+k}))
\end{equation}

\begin{proof}

\begin{align*}
    &\mathbb{P}_{\pi,\mu}(a \mid x',x)\\ &\stackrel{(a)}{=} \frac{\mathbb{P}_{\pi,\mu}(x' \mid x,a)\mathbb{P}_{\pi,\mu}(a\mid x)}{\sum_{a'} \mathbb{P}_{\pi,\mu}(x' \mid x,a')}\\
    &\stackrel{(b)}{=} \frac{\mathbb{P}_{\pi,\mu}(x' \mid x,a)\pi(a\mid \phi_\star(x)))}{\sum_{a'} \mathbb{P}_{\pi,\mu}(x' \mid x,a')\pi(a'\mid \phi_\star(x))}\\
    &\stackrel{(c)}{=} \frac{q(x' \mid \phi_\star(x'),\phi_{\star,e}(x')) \mathbb{P}_{\pi,\mu}(\phi_\star(x')\mid \phi_\star(x),a) \mathbb{P}_{\pi,\mu}(\phi_{\star,e}(x') \mid \phi_{\star,e}(x))\pi(a\mid \phi_\star(x))}{\sum_{a'} q(x' \mid \phi_\star(x'),\phi_{\star,e}(x')) \mathbb{P}_{\pi,\mu}(\phi_\star(x')\mid \phi_\star(x),a') \mathbb{P}_{\pi,\mu}(\phi_{\star,e}(x') \mid \phi_{\star,e}(x))\pi(a'\mid \phi_\star(x))}\\
    &=\frac{\mathbb{P}_{\pi,\mu}(\phi_\star(x')\mid \phi_\star(x),a)\pi(a\mid \phi_\star(x))}{\sum_{a'} \mathbb{P}_{\pi,\mu}(\phi_\star(x')\mid \phi_\star(x),a')\pi(a'\mid \phi_\star(x))}.
\end{align*}

\end{proof}

Relation $(a)$ holds by Bayes' theorem. Relation $(b)$ holds by the assumption that $\pi$ is uniformly random (in the first proof) or exo-free (in the second proof). Relation $(c)$ holds by the factorization property.  Thus, $\mathbb{P}_{\pi,\mu}(a \mid x',x)= \mathbb{P}_{\pi,\mu}(a \mid \phi_\star(x'),\phi_\star(x)),$ and is constant upon changing the observation while fixing the agent-centric state.

\subsection{Connection Between ACRO and Behavior Cloning}
\label{app:bceq}

In the special case where all data is collected under a fixed deterministic, exogenous-free policy, \methodname and behavior cloning become equivalent.  This case can still be non-trivial, if the start state of the episode is stochastic or if the environment dynamics are stochastic.  

\begin{lemma} Under fixed, deterministic, and exo-free policy $\hat{\pi}: \Xcal \rightarrow \Acal$, multi-step inverse model is equivalent to behavior cloning.  For all $a_t \in \Acal$, and $(x_t, x_{t+k})$ such that $ \PP_{\hat{\pi}}(X_t=x_t, X_{t+k}=x_{t+k})>0$ we have:
\begin{equation}
    \PP_{\hat{\pi}}(a_t \mid \phi_\star(x_t), \phi_\star(x_{t+k})) = \PP_{\hat{\pi}}(a_t \mid \phi_\star(x_t))
\end{equation}
\label{lemma:bceq}
\end{lemma}

The proof of this claim is simply that behavior cloning is already able to predict actions perfectly in this special case, so there can be no benefit to conditioning on future observations.  

\begin{proof}

By the assumption of the deterministic exo-free policy, we have that $\PP_{\hat{\pi}}(a_t = \hat{a}(\phi_\star(x_t)) \mid \phi_\star(x_t)) = 1$ where $\hat{a} : \mathcal{S} \xrightarrow[]{} A$ is a function mapping the latent state to the action.  

Using bayes theorem we write: 

\begin{align}
\PP_{\hat{\pi}}(a_t \mid \phi_\star(x_t), \phi_\star(x_{t+k})) =
\frac{\PP_{\hat{\pi}}(\phi_\star(x_{t+k} \mid \phi_\star(x_t), a_t) \PP_{\hat{\pi}}(a_t \mid \phi_\star(x_t))}{\sum_{a''} \PP_{\hat{\pi}}(\phi_\star(x_{t+k} \mid \phi_\star(x_t), a'') \PP_{\hat{\pi}}(a'' \mid \phi_\star(x_t))}
\end{align}

For any examples in the dataset and for all $a' \in A$: 

Case 1: $a' = \hat{a}(\phi_\star(x_t))$. It holds that 

\begin{align*}
\PP_{\hat{\pi}}(a_t \mid \phi_\star(x_t), \phi_\star(x_{t+k})) &= \frac{\PP_{\hat{\pi}}(\phi_\star(x_{t+k} \mid \phi_\star(x_t), a_t) \PP_{\hat{\pi}}(a_t \mid \phi_\star(x_t))}{\sum_{a''} \PP_{\hat{\pi}}(\phi_\star(x_{t+k} \mid \phi_\star(x_t), a'') \PP_{\hat{\pi}}(a'' \mid \phi_\star(x_t))}\\
\PP_{\hat{\pi}}(a_t \mid \phi_\star(x_t), \phi_\star(x_{t+k})) &= \frac{\PP_{\hat{\pi}}(\phi_\star(x_{t+k} \mid \phi_\star(x_t), a_t = a')}{\PP_{\hat{\pi}}(\phi_\star(x_{t+k} \mid \phi_\star(x_t), a_t = a')}\\
\PP_{\hat{\pi}}(a_t \mid \phi_\star(x_t), \phi_\star(x_{t+k})) &= 1.
\end{align*}

On the other hand, it holds that $\PP_{\hat{\pi}}(a_t = a' \mid \phi_\star(x_t)) = 1$ since $a' = \hat{a}(\phi_\star(x_t))$. Hence, for this case, the claim holds true.

Case 2: $a' \neq \hat{a}(\phi_\star(x_t))$. It holds that 

\begin{align*}
\PP_{\hat{\pi}}(a_t \mid \phi_\star(x_t), \phi_\star(x_{t+k})) &= \frac{\PP_{\hat{\pi}}(\phi_\star(x_{t+k} \mid \phi_\star(x_t), a_t) \PP_{\hat{\pi}}(a_t \mid \phi_\star(x_t))}{\sum_{a''} \PP_{\hat{\pi}}(\phi_\star(x_{t+k} \mid \phi_\star(x_t), a'') \PP_{\hat{\pi}}(a'' \mid \phi_\star(x_t))}\\
\PP_{\hat{\pi}}(a_t \mid \phi_\star(x_t), \phi_\star(x_{t+k})) &= \frac{0}{\PP_{\hat{\pi}}(\phi_\star(x_{t+k} \mid \phi_\star(x_t), a_t = \hat{a}(\phi_\star(x_t)))}\\
\PP_{\hat{\pi}}(a_t \mid \phi_\star(x_t), \phi_\star(x_{t+k})) &= 0.
\end{align*}
On the other hand, it holds that $\PP_{\hat{\pi}}(a_t = a' \mid \phi_\star(x_t)) = 0$ since $a' \neq \hat{a}(\phi_\star(x_t))$. Hence, for this case the claim also holds true. This concludes the proof since the two distributions are equal in both cases.  
\end{proof}

\subsection{Predicting the First Action vs. Predicting Action Sequences}
\label{app:acro_sequence_action_prediction}

In \methodname we only predict the first action from $x_t$ to $x_{t+k}$ rather than predicting the entire action sequence.  In an environment with deterministic dynamics, we will prove that these two approaches are asymptotically equivalent.  For the proof we will also make a stronger assumption that the dynamics are deterministic in the learned latent space, i.e. for the learned encoder $\phi$, there exists a function $f$ such that: $\phi(x_{j}) = f(\phi(x_t),  a_{t:j})$.  This assumption will hold for the optimal $\phi$, and it is also likely to be empirically true since $\phi$ is a high-dimensional continuous latent state, thus no two points are likely to have exactly the same representation.  In a stochastic environment, the two approaches are different, but we will provide a counter-example to make the case against predicting action sequences.  

The ACRO objective optimizes the following 
\begin{align}
\phi_\star \in \argmax_{\phi \in \Phi} \E_{t \sim U(0,N)} \E_{k \sim U(0,K)} \text{log}\left(\PP(a_t \mid \phi(x_t), \phi(x_{t+k}))\right). \label{eqn:acro}
\end{align}
The $k$ step action sequence prediction approach optimizes: 
\begin{equation}
\label{eq:glamor_objective}
\phi_\star \in \argmax_{\phi \in \Phi} \E_{t \sim U(0,N)} \E_{k \sim U(0,K)}  \text{log}\left(\PP(a_t, \dots a_{t+k} \mid \phi(x_t), \phi(x_{t+k}))\right).
\end{equation}

\subsubsection{Deterministic Dynamics}

\begin{align}
    \PP(a_{t:t+k} \mid \phi(x_t), \phi(x_{t+k})) &= \prod_{j=t}^{t+k} \PP(a_j \mid \phi(x_t), \phi(x_{t+k}), a_{t:j}) 
\end{align}

By the assumption of deterministic dynamics in the latent space: 

\begin{align}
    \PP(a_{t:t+k} \mid \phi(x_t), \phi(x_{t+k})) &= \prod_{j=t}^{t+k} \PP(a_j \mid \phi(x_t), a_{t:j}, \phi(x_{t+k})) 
\end{align}

After applying the markov assumption as we have assumed an MDP: 

\begin{align}
    \PP(a_{t:t+k} \mid \phi(x_t), \phi(x_{t+k})) &= \prod_{j=t}^{t+k} \PP(a_j \mid \phi(x_j), \phi(x_{t+k})) 
\end{align}

Now we can put this back into the k-step action sequence prediction problem: 

\begin{align}
\phi_\star &\in \argmax_{\phi \in \Phi} \E_{t \sim U(0,N)} \E_{k \sim U(0,K)}  \text{log}\left(\prod_{j=t}^{t+k} \PP(a_j \mid \phi(x_j), \phi(x_{t+k}))\right) \\
\phi_\star &\in \argmax_{\phi \in \Phi} \E_{t \sim U(0,N)} \E_{k \sim U(0,K)} \sum_{j=t}^{t+k} \text{log}\left(\PP(a_j \mid \phi(x_j), \phi(x_{t+k}))\right) \\
\phi_\star &\in \argmax_{\phi \in \Phi} \E_{t \sim U(0,N)} \E_{k \sim U(0,K)} \E_{j \sim U(t,t+k)} \text{log}\left(\PP(a_j \mid \phi(x_j), \phi(x_{t+k}))\right)
\end{align}

When $N \gg k$, the distributions of $t$ and $j$ converge, and we can write: 

\begin{align}
\phi_\star &\in \argmax_{\phi \in \Phi} \E_{t \sim U(0,N)} \E_{k \sim U(0,K)} \text{log}\left(\PP(a_t \mid \phi(x_t), \phi(x_{t+k}))\right)
\end{align}

which we can see is the same as the first-action prediction objective that \methodname optimizes.  

\subsubsection{Stochastic Dynamics}

If the environment has stochastic dynamics, predicting future actions to reach a goal state without conditioning on the preceding observations is very difficult.  This is because what action needs to be taken may depend on what actually happened in the environment.  For example, if I'm playing a video game, and there is a small chance that the game pauses for one minute, the actions will need to depend on whether the pause occurred.  We can construct a stochastic environment in which every action following the first action is completely unpredictable.  We can imagine an environment where there is some information in the observation space which is set randomly on every step and indicates how the agent's controls are randomly permuted for that step.  In principle, the first-action predictor can easily use this information to adapt what actions it predicts, and can obtain its original accuracy given sufficient model capacity.  On the other hand, the action-sequence predictor  $\left(\PP(a_t, \dots a_{t+k} \mid \phi(x_t), \phi(x_{t+k}))\right)$ will only be able to predict the first action well, and can have no better than uniformly random accuracy at predicting the remaining actions.  This is because only $\phi(x_t)$ contains the information about what has happened in the environment which is necessary for control, the history of past actions do not contain the necessary information.  In this example, predicting the sequence of actions makes the task much noisier, while providing no additional signal for the model.  

\subsection{Counterexample for One-Step Inverse Models}
\label{sec:counterexample}
\begin{figure}
    \begin{center}
        \scalebox{1}{
                \begin{tikzpicture}[->, >=stealth', scale=0.6 , semithick, node distance=2cm]
                    \tikzstyle{every state}=[fill=white,draw=black,thick,text=black,scale=1]
                    \node[state]    (x0)                     {$s_0$};
                    \node[state]    (x1)[right of=x0]   {$s_1$};
                    \node[state]    (x2)[right of=x1]   {$s_2$};
                    \node[state]    (x3)[right of=x2]   {$s_3$};
                    \node[state,accepting]    (xf)[below of=x0]   {$s_f$};
                    \path
                    (x0) edge[above]    node{$a_1$}     (x1)
                    (x1) edge[above]    node{$a_1$}     (x2)
                    (x2) edge[above]    node{$a_1$}     (x3)
                    (x0) edge[right]    node{$a_2$}     (xf)
                    (x1) edge[right]    node{$a_2$}     (xf)
                    (x2) edge[right, line width=0.5mm]    node{$a_2$}     (xf)
                    (x3) edge[right]    node{$a_2$}     (xf)
                    (x0) edge[out=150, in=210, loop, right]    node{$a_0$}     (x0)
                    (x1) edge[out=60, in=120, loop, below]    node{$a_0$}     (x1)
                    (x2) edge[out=60, in=120, loop, below]    node{$a_0$}     (x2)
                    (x3) edge[out=330, in=30, loop, left]    node{$a_0$}     (x3)
                    (x3) edge[out=45, in=135, above]    node{$a_1$}     (x0);
                \end{tikzpicture}
        }
        \caption{An MDP where one-step models can fail.}
        \label{fig:minimalMDP}
    \end{center}
\end{figure}
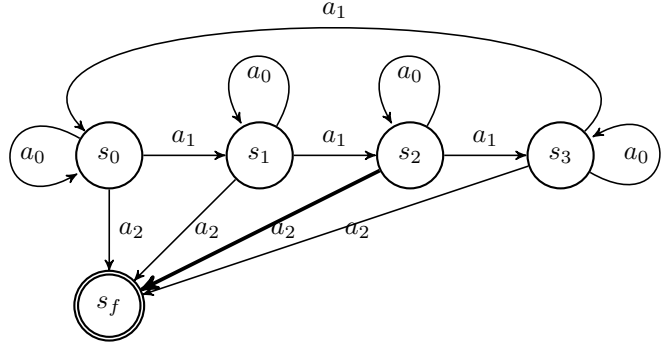

In this section, we build an MDP and a dataset of trajectories in that MDP where a one-step model will fail to allow the learning of the optimal policy. The MDP can be seen in Figure~\ref{fig:minimalMDP}, all transitions are deterministic, and yield $0$ reward, apart for action $a_2$ in state $s_2$, which gives a reward of 1 (bold arrow in the graph). The final state $s_f$ denotes the termination of the trajectory

We consider the dataset comprised of the following five trajectories, formatted as $\langle s_t,a_t,r_t,s_{t+1} \dots \rangle$:
\begin{align}
    \mathcal{D}=\{&\underbrace{\langle s_0,a_0,0,s_0,a_0,0,s_0,a_2,0,s_f\rangle}_{\tau_1}, \\
    &\underbrace{\langle s_0,a_1,0,s_1,a_0,0,s_1,a_2,0,s_f\rangle}_{\tau_2}, \\
    &\underbrace{\langle s_0,a_1,0,s_1,a_1,0,s_2,a_0,0,s_2,a_2,\textbf{1},s_f\rangle}_{\tau_3}, \\
    &\underbrace{\langle s_0,a_1,0,s_1,a_1,0,s_2,a_1,0,s_3,a_0,0,s_3,a_2,0,s_f\rangle}_{\tau_4}, \\
    &\underbrace{\langle s_0,a_1,0,s_1,a_1,0,s_2,a_1,0,s_3,a_1,0,s_0,a_2,0,s_f\rangle}_{\tau_5} \}.
\end{align}
$\tau_1$ loops twice in $s_0$ and then terminates. The other four trajectories reach $s_1$, $s_2$, $s_3$ and $s_0$ by taking $a_1$ a minimal number of times, then loop once (except $\tau_5$), and terminate. We note that this dataset covers the full state and action space.

It is possible to reach a 0 loss with a one-step inverse model $p_1$ built on top of the following representation $\phi$ from $\mathcal{S}$ to $\mathcal{X} = \{x_0, x_1, x_f\}$: $\phi(s_0) = \phi(s_2) = x_0$, $\phi(s_1) = \phi(s_3) = x_1$ and $\phi(s_f) = x_f$:
\begin{align}
    p_1(a | x_0, x_0) &= \delta_{a = a_0}, \\
    p_1(a | x_0, x_1) &= \delta_{a = a_1}, \\
    p_1(a | x_0, x_f) &= \delta_{a = a_2}.
\end{align}

Now, once projected onto $\mathcal{X}$, the dataset becomes:

\begin{align}
    \mathcal{D}=\{&\underbrace{\langle x_0,a_0,0,x_0,a_0,0,x_0,a_2,0,x_f\rangle}_{\tau_1}, \\
    &\underbrace{\langle x_0,a_1,0,x_1,a_0,0,x_1,a_2,0,x_f\rangle}_{\tau_2}, \\
    &\underbrace{\langle x_0,a_1,0,x_1,a_1,0,x_0,a_0,0,x_0,a_2,\textbf{1},x_f\rangle}_{\tau_3}, \\
    &\underbrace{\langle x_0,a_1,0,x_1,a_1,0,x_0,a_1,0,x_1,a_0,0,x_1,a_2,0,x_f\rangle}_{\tau_4}, \\
    &\underbrace{\langle x_0,a_1,0,x_1,a_1,0,x_0,a_1,0,x_1,a_1,0,x_0,a_2,0,x_f\rangle}_{\tau_5} \}.
\end{align}

We see that action $a_2$ in state $x_0$ has an expected reward of $1 / 3$, and it is the only state-action pair with a non-zero expected reward. Any reasonable offline RL algorithm applied to this dataset will output a policy with a non-zero probability assigned to that action in $x_0$. However, executing that policy, \textit{i.e.,} taking action $a_2$ in state $s_0$ terminates the episode with 0 reward, which is suboptimal. 

On the other hand, an optimal two-step model (and by extension an n-step one) will not collapse states $s_0$ and $s_2$ as this would prevent distinguishing the first action taken in $\tau_1$ from the first action taken in $\tau_3$ (corresponding respectively to pairs $(s_0, s_0)$ and $(s_0, s_2)$ after two timesteps). Consequently, an offline RL applied in representation can still learn an optimal policy.

\section{Extended Related Work}
\label{app:related_work}
\textbf{Representation learning in Offline RL}. 
Representation learning offers an exciting avenue to address the demands of learning compact feature for state by incorporating the auxiliary task of the state feature within the learning task. Empirical studies on representation learning in Offline RL have been first addressed by \citet{DBLP:conf/icml/YangN21}, which evaluate the ability of a broad set of representation learning objectives in the offline dataset and propose Attentive Contrastive Learning (ACL) to improve downstream policy performance. After that, ~\citet{DBLP:conf/nips/ChenCTWEFLAWL0A21} investigate whether the auxiliary representation learning objectives that broadly used in NLP or CV domains can help for imitation across different Offline RL tasks. ~\citet{lu2022challenges} further explores the existing challenges for visual observation input with the Offline RL dataset, meanwhile providing simple modifications on several state-of-the-art Offline RL algorithms to establish a competitive baseline. ~\citet{zang2023behavior} leverages behavior cloning for state representation learning to improve the downstream Offline policy performance.  Another branch of representation learning in Offline RL is theoretical side.~\citet{DBLP:journals/corr/abs-2110-04652} studies the representation learning in low-rank MDPs with Offline settings and proposes an algorithm that leverages pessimism to learn under a partial coverage condition, ~\citet{DBLP:conf/nips/NachumY21} develops a representation objective that provably accelerate the sample-efficiency of downstream Offline RL tasks, ~\citet{DBLP:conf/icml/GhoshB20} theoretically shows that the stability of the policy is tightly connected with the geometry of the transition matrix, which can provide stability conditions for algorithms that learn features from the transition matrix of a policy and rewards. 

\textbf{Reward-Dependent Methods}. \textsc{DrQV2}~\citep{kostrikov2020image} learns a value function from offline tuples of observations, rewards, and actions. This could feasibly ignore exogenous noise given a suitable data-collection policy, but will not generally learn the full agent-centric latent state due to a heavy dependence on the reward structure. Bisimulation-based methods ( \textsc{DBC}~\cite{zhang2020learning}, \textsc{MICo}~\cite{mico}, and ~\textsc{SimSR}~\cite{simsr}) learns representations which have similar values under a learned value function. In general, bisimulation is an overly restrictive state abstraction that fails to transfer to different tasks.

\textbf{Offline RL}. The predominant approach to train offline RL agent is regularizing the learned policy to be close to the behavior policy of the offline dataset. This can be implemented by generating the actions that similar to the dataset and restricting the output of the learned policy close to the generated actions~\citep{bcq}, penalizing the distance between the learned policy and the behavior of the dataset~\citep{KumarFSTL19, DBLP:conf/acml/ZhangKP21}, or introducing a pessimism term to regularize the Q function for avoiding high Q value of the out-of-distribution actions~\citep{DBLP:conf/nips/KumarZTL20,DBLP:conf/iclr/BuckmanGB21}. Some approaches utilize BC as a reference for policy optimization with the baseline methods~\citep{DBLP:conf/nips/FujimotoG21, DBLP:conf/icml/LarocheTC19,Nadjahi2019,Simao2020,DBLP:conf/rss/RajeswaranKGVST18}. Some other approaches improve the performance by measuring the uncertainty of the model's prediction~\citep{DBLP:conf/nips/YuTYEZLFM20, DBLP:conf/nips/KidambiRNJ20, DBLP:conf/nips/AnMKS21}.

\section{Additional Experiment Results}
\label{app:additional_experiments}

\subsection{Manipulation Gridworld Environments}
\label{app:blockworld}

We investigated the ability of \methodname on whether it removes task relevant information as well or not. For this, we constructed a simple manipulation task on gridworlds, as described earlier in the paper. Detailed results are provided in Table~\ref{tab:blockworld}.  

\begin{table}[h!]
    \centering
        \caption{Probing Accuracy on the position of the agent, the position of the block, and the exogenous noise after training with \methodname.  We found that the position of the agent and the block are both captured by the representation, while exogenous noise is discarded.  }
    \label{tab:blockworld}

    \begin{tabular}{c c | c c c}
         \textbf{Task} & \textbf{Size} & \textbf{Agent Position} & \textbf{Block Position} & \textbf{Exogenous Noise} \\
         \shline
         Push & 4x4 & 100.00 & 100.00 & 6.239 \\
         Push & 5x5 & 100.00 & 100.00 & 3.991 \\
         Push & 6x6 & 100.00 & 100.00 & 2.752 \\
         Push & 7x7 & 97.27  & 100.00 & 1.982 \\
         Push & 8x8 & 94.53  & 99.61 & 1.621 \\
         \hline
         Pull & 4x4 & 100.00 & 100.00 & 6.221 \\
         Pull & 5x5 & 100.00 & 100.00 & 4.000 \\
         Pull & 6x6 & 100.00 & 100.00 & 2.816 \\
         Pull & 7x7 & 100.00 & 100.00 & 2.048 \\
         Pull & 8x8 & 100.00 & 100.00 & 1.566 \\
    \end{tabular}
\end{table}

\subsection{Atari Experiments with Exogenous Information} 
\label{sec:atari-results}
\begin{figure}[h]
    \centering
\includegraphics[width=0.30\textwidth, trim  = {0 10cm 0 10cm}, clip]{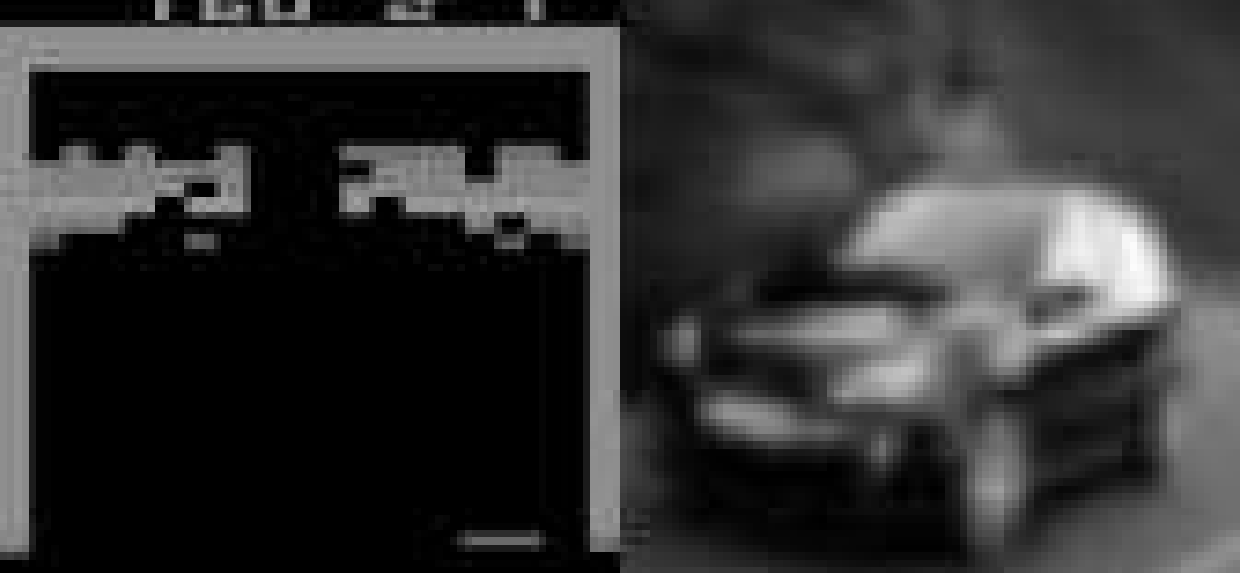}
    \caption{\textbf{Example} of an observation from Breakout with a CIFAR10 image on the side.}
    \label{fig:atari_example}
\end{figure}

We also consider the setting of Atari. We build on the setup introduced in decision transformers \citep{NEURIPS2021DT}. While decision transformers focus on framing the reinforcement learning problem as a sequence modeling problem, we mainly focus on learning representations which can learn to ignore exogenous noise. As such, for the decision transformer, we use trajectories of latent state and action pairs, instead of raw states as input, for modelling the sequence of actions. We use the same 4 games used in \cite{NEURIPS2021DT} - Pong, Qbert, Breakout, and Seaquest. We consider the \mediumexo setting where a different image is used in each episode and concatenated to the side of each observation. We add randomly sampled images from the CIFAR10 dataset \citep{Krizhevsky_2009_17719} as exogenous noise.~\pref{fig:atari_example}  shows an example observation from breakout with exogenous noise.

Decision Transformers use the DQN-Replay dataset \citep{agarwal2020optimistic} for training. The model is trained using a sequence modeling objective to predict the next action given the past states, actions, and returns-to-go $\hat{R}_c = \sum_{c'=c}^C r_c$, where $c$ denotes the timesteps. This results in the following trajectory representation: $\tau = \big( \hat{R}_1, s_1, a_1, \hat{R}_2, s_2, a_2, \hat{R}_3, s_3, a_3, \ldots \big)$, where $a_c$ denotes the actions and $s_c$ denotes the states.  At test time, the start state $s_1$ and desired return $\hat{R}_1$ is fed into the model and it autoregressively generates the rest of the trajectory.

Decision Transformer uses a convolutional encoder to encode the observations. We first pretrain this encoder using the proposed $\methodname$ objective. We use the 1-step inverse objective and DRIML as our baselines. After pretraining, we train the decision transformer using the sequence modeling objective keeping the encoder fixed. We present results in~\pref{tab:atari_results}. We can see that ACRO outperforms both the baselines in all games further showing the effectiveness of the proposed approach. 

\textbf{Hyperparameter Details}. We keep most of the hyperparameter details same as used in \citet{NEURIPS2021DT}. They use episodes of fixed length during training - also referred to as the \textit{context length}. We use a context length of 30 for Seaquest and Breakout and 50 for Pong and Qbert. Similar to \citet{NEURIPS2021DT}, we consider one observation to be a stack of 4 atari frames. To implement the ACRO objective, we sample 8 different values for $k$ and calculate the objective for each value of $k$, obtaining the final loss by taking the sum across all the sampled values of $k$. We do not feed the embedding for $k$ in the MLP that predicts the action while computing the ACRO objective.

\begin{table}[h]
    \centering
        \caption{\textbf{Atari (\mediumexo)}. Here we compare ACRO to One-Step-Inverse model and DRIML on various games from the Atari benchmark with exo-noise. We can see that ACRO outperforms the baselines in all but one case. Results averaged across 5 seeds. }
    \label{tab:atari_results}
    \resizebox{0.6\columnwidth}{!}{%
    \begin{tabular}{c | c c c c}
         \textbf{Game} & \textbf{1-Step Inv.} & \textbf{DRIML} & \textbf{AC-State} & \textbf{ACRO} \\
         \shline
         Breakout & \g{3.8}{0.4}& \g{1.0}{0.0} & \g{19.4}{3.323}&  \highlight{\g{20.6}{3.2}} \\
         Pong & \g{8.6}{3.2} & \g{-20.0}{0.0} & \g{9.0}{3.578 }& \highlight{\g{11.8}{3.37}} \\
         Qbert & \g{536.2}{233.75} & \g{277.8}{46.24} & \g{388.4}{184.22} & \highlight{\g{657.4}{271.52}} \\
         Seaquest & \g{274.0}{29.61} & \g{94.4}{4.63} & \highlight{\g{984.8}{80.31}} &\g{972.4}{136.09} \\
    \end{tabular}
    }
\end{table}

\subsection{Experimental Comparison between Single Action vs Multiple Actions Prediction}
\label{app:rebuttal_generic_multi_action_prediction}
\begin{figure*}[!htb]
\centering
\centering
\subfigure[Halfcheetah-expert]{
\includegraphics[width=0.31\textwidth]{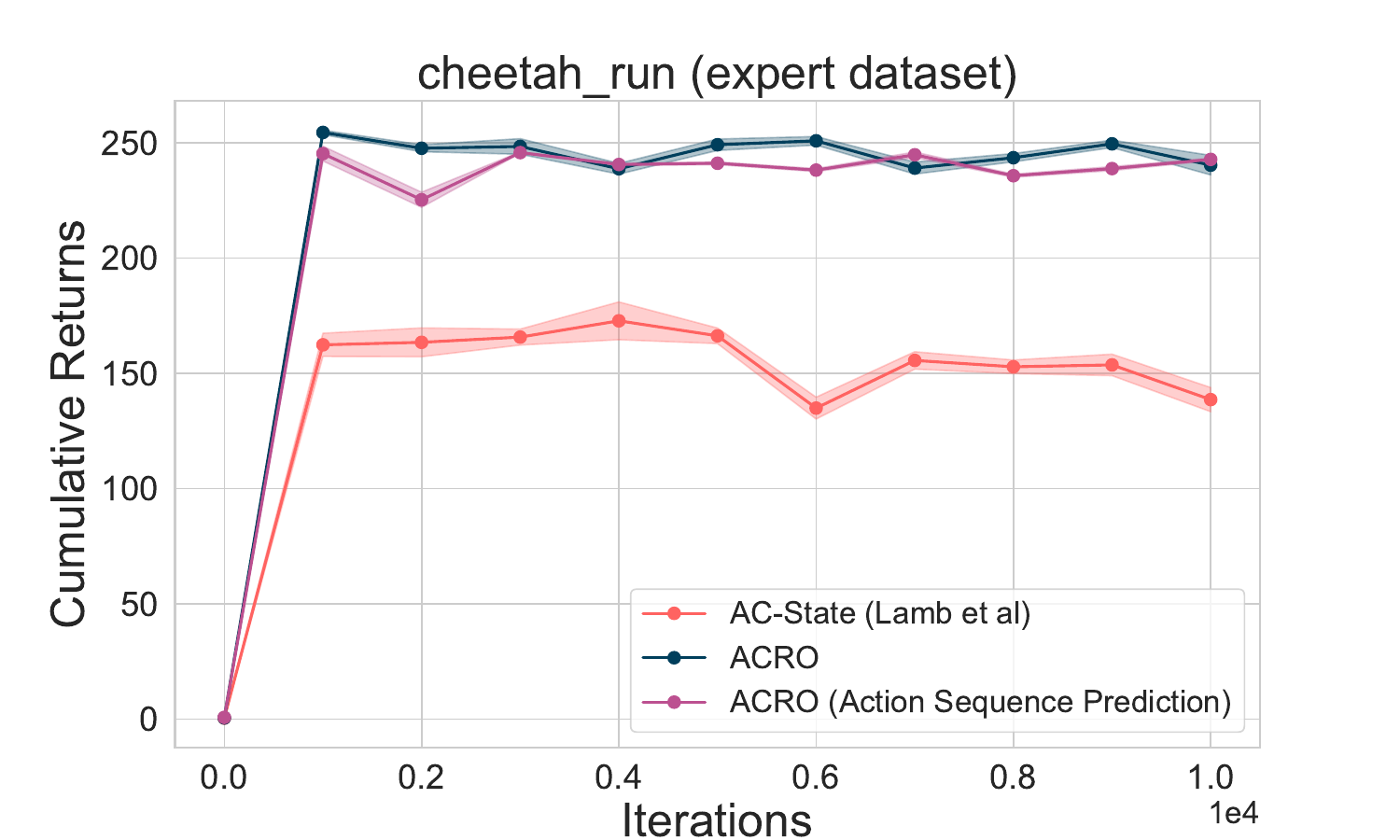}
}
\subfigure[Humanoid-Walk-expert]{
\includegraphics[width=0.31\textwidth]{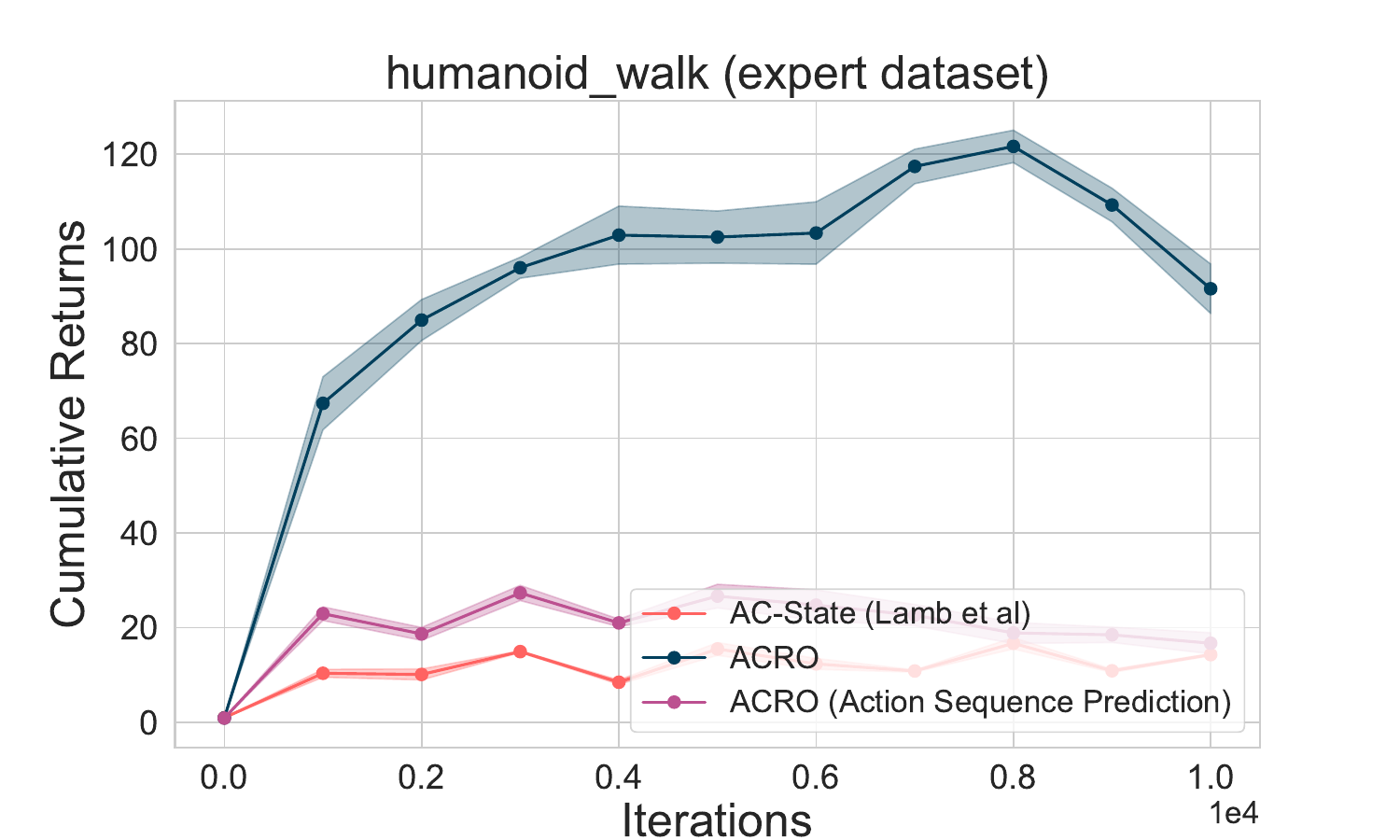}
}
\subfigure[Walker-Walk-expert]{
\includegraphics[width=0.31\textwidth]{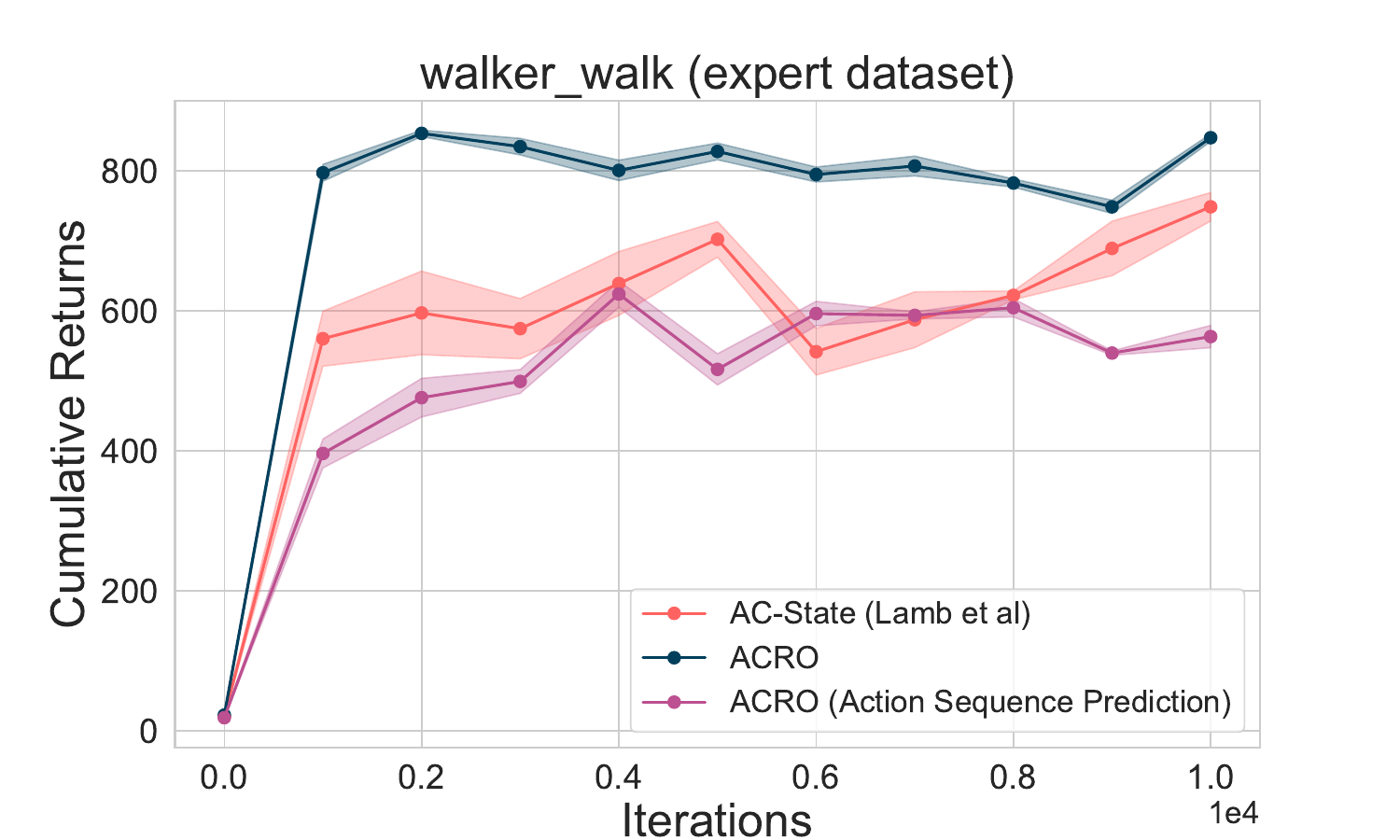}
}
\caption{Comparions between \methodname, \methodname with multiple action predicton (ie, predicting an action sequence) and also with AC-State \citep{lamb2022guaranteed} in \textbf{time-correlated} \hardexo offline datasets}
\label{fig:single_multiple_action_pred_comparison_time_correlated}
\end{figure*}

\begin{figure*}[!htb]
\centering
\centering
\subfigure[Halfcheetah-expert]{
\includegraphics[width=0.31\textwidth]{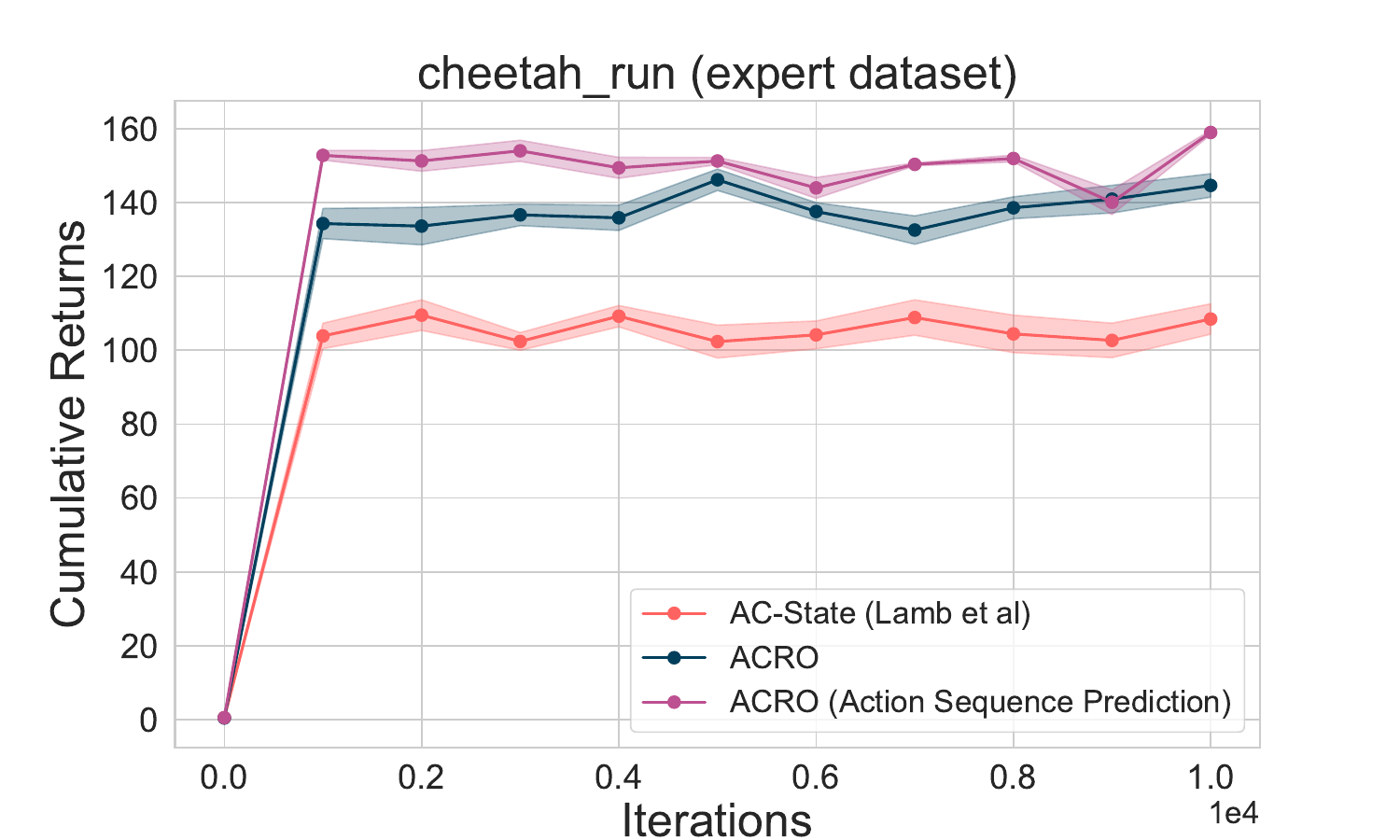}
}
\subfigure[Humanoid-Walk-expert]{
\includegraphics[width=0.31\textwidth]{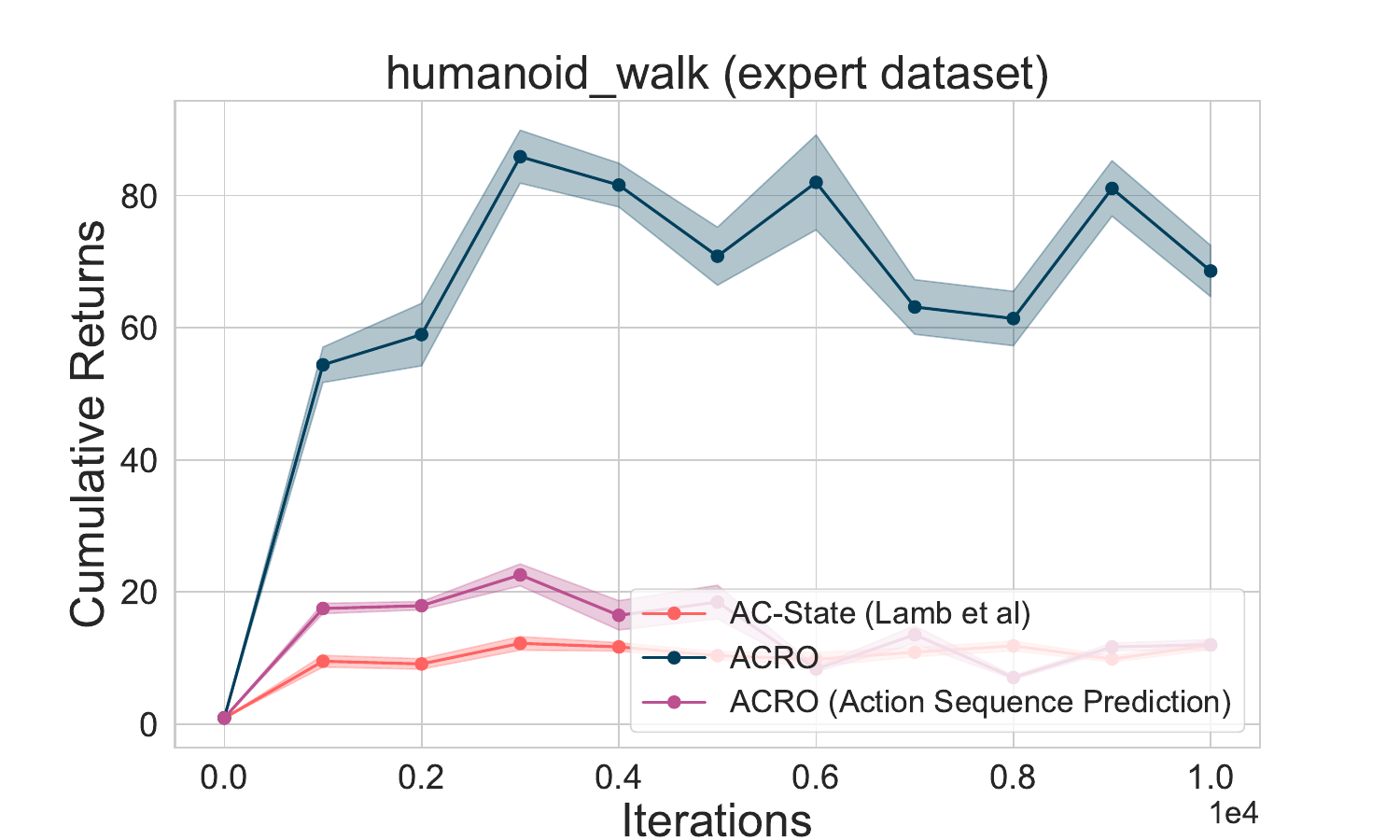}
}
\subfigure[Walker-Walk-expert]{
\includegraphics[width=0.31\textwidth]{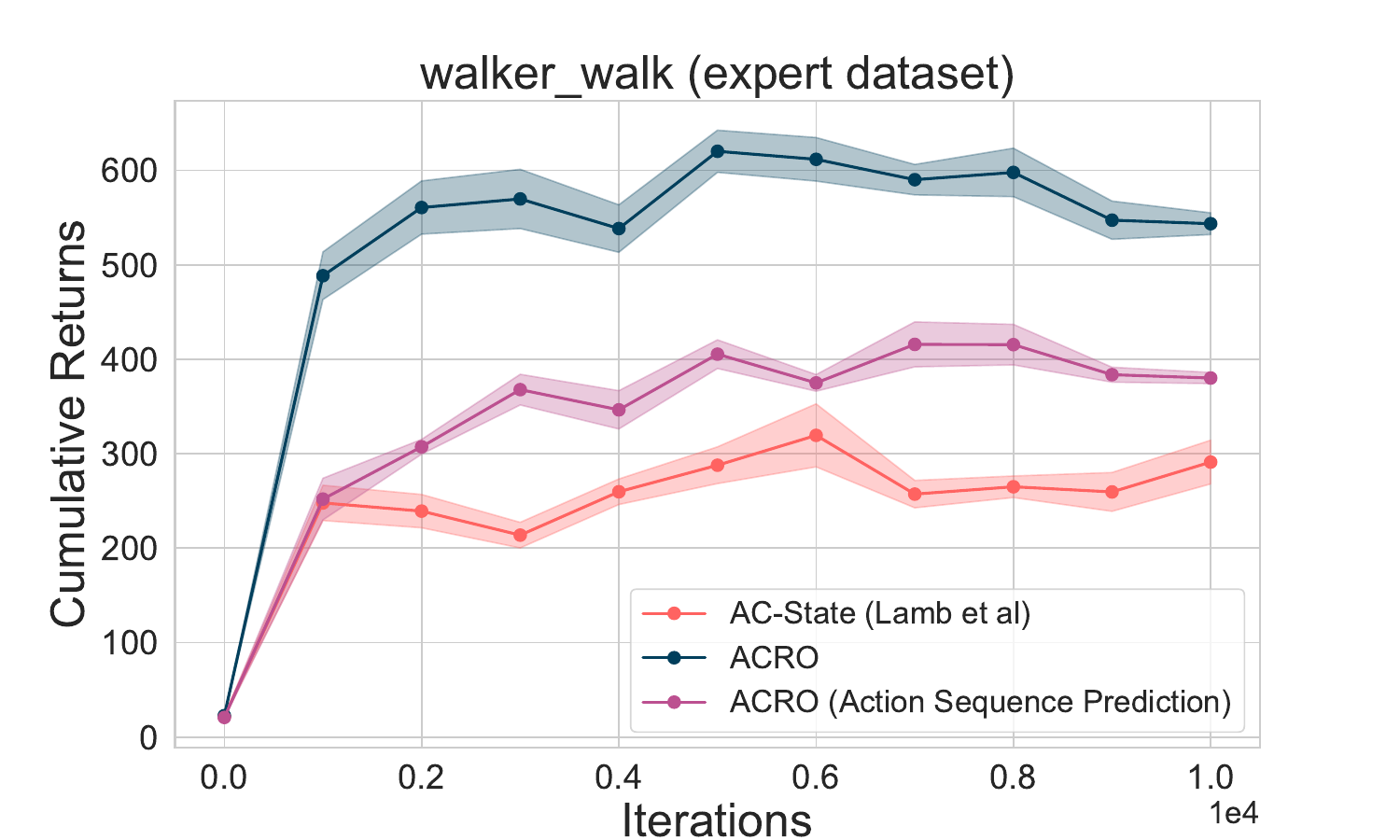}
}
\caption{Comparions between \methodname, \methodname with multiple action predicton (ie, predicting an action sequence) and also with AC-State \citep{lamb2022guaranteed} in \textbf{changing video} \hardexo offline datasets}
\label{fig:single_multiple_action_pred_comparison_change_video}
\end{figure*}

We empirically investigate how the \methodname algorithm compares if we use multi-action prediction up to the k-th timestep, in comparison to only predicting a single next step action.  Since \methodname already conditions on $\phi(x_{t+k})$ for k-th step in the future, it is natural to ask how the performance varies if we predict multiple future actions compared to a single action. Concretely, the ACRO objective optimizes the following 
\begin{align}
\phi_\star \in \argmax_{\phi \in \Phi} \E_{t \sim U(0,N)} \E_{k \sim U(0,K)}  \text{log}\left(\PP(a_t \mid \phi(x_t), \phi(x_{t+k}))\right). \label{eqn:acro}
\end{align}
whereas, we could instead predict the $k$ step action sequence: 
\begin{equation}
\label{eq:glamor_objective}
\phi_\star \in \argmax_{\phi \in \Phi} \E_{t \sim U(0,N)} \E_{k \sim U(0,K)}  \text{log}\left(\PP(a_t, \dots a_{t+k} \mid \phi(x_t), \phi(x_{t+k}))\right).
\end{equation}
where equation \ref{eq:glamor_objective} is implemented using an LSTM that outputs the $k$ actions. One reason to prefer predicting just the first action is that it is a simpler model and is computationally cheaper (as it requires just a classifier over actions and not an autoregressive model over sequences like the LSTM).  Intuitively, we also felt that predicting the first action to reach a goal would be sufficient, because ultimately every action along the trajectory is still predicted, but conditioned on the observation prior to the action being taken.  In an environment with stochastic dynamics, we see this as being better in principle, because the best action to take at a given step is dependent on what has happened in the environment.  In a deterministic environment, both approaches are valid in principle. Nonetheless, we agree that it is important to also answer this question experimentally.  Figures \ref{fig:single_multiple_action_pred_comparison_time_correlated} and \ref{fig:single_multiple_action_pred_comparison_change_video} show comparison between $\methodname$ and a variation of $\methodname$ predicting multiple actions in the future. We use the same training setup, where all the representations are pre-trained in presence of \hardexo noise in observations.

In addition, we also compare \methodname with the AC-State objective \cite{lamb2022guaranteed} which requires an additional information bottleneck based auxilliary objective for removing exogenous information. \cite{lamb2022guaranteed} uses a vector quantization bottleneck \cite{vqvae} in addition to a multi-step inverse dynamics objective for removal of exogenous noise. In contrast, \methodname is much more straightforward, without requiring any fine-tuning with information bottlenecks, and we demonstrate the significance of \methodname compared to both these objectives, as shown in figures \ref{fig:single_multiple_action_pred_comparison_time_correlated} and \ref{fig:single_multiple_action_pred_comparison_change_video}

\subsection{Data coverage and role of exploration in offline RL}
\label{app:data_coverage}

In this section, we investigate the ability of learnt representations from \methodname and other baselines, depending on the dataset coverage in the offline setting. This leads to interesting insights on the role of data and exploration in offline RL, and how the quality of learnt representation depends on coverage data, for learning robust representations. From theoretical insights studying offline RL, it is clear that lack of dataeset coverage would degrade the algorithm. The fact that the concentrability coefficient is finite implies there is a good dataset coverage.  In this work, we mostly focused on this problem from its empirical side and established the failure of existing approaches, as well as offering a fix (that works under the dataset coverage assumption). Further, the online RL problem with exogenous noise is also of interest. As of now, there is no algorithm with provable guarantees for the general RL problem with exogenous noise (there are some solutions under different sets of assumptions as we elaborated on in the related work section). Our work does not tackle this challenging problem, but study the offline aspects of its; a prevalent problem from a practical perspective.

\begin{figure*}[!htb]
\centering
\centering
\subfigure[\methodname (ours)]{
\includegraphics[width=0.45\textwidth]{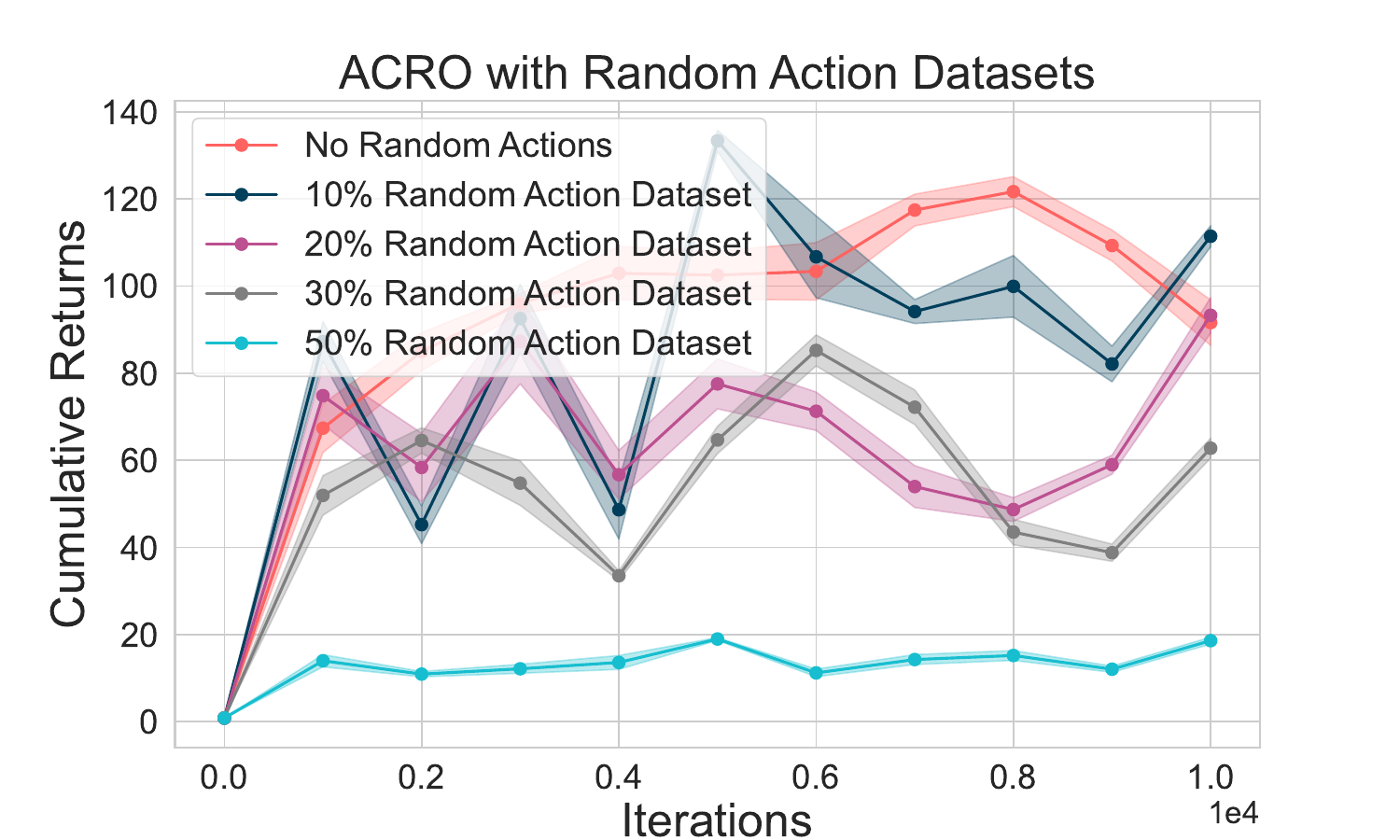}
}
\subfigure[AC-State \cite{lamb2022guaranteed}]{
\includegraphics[width=0.45\textwidth]{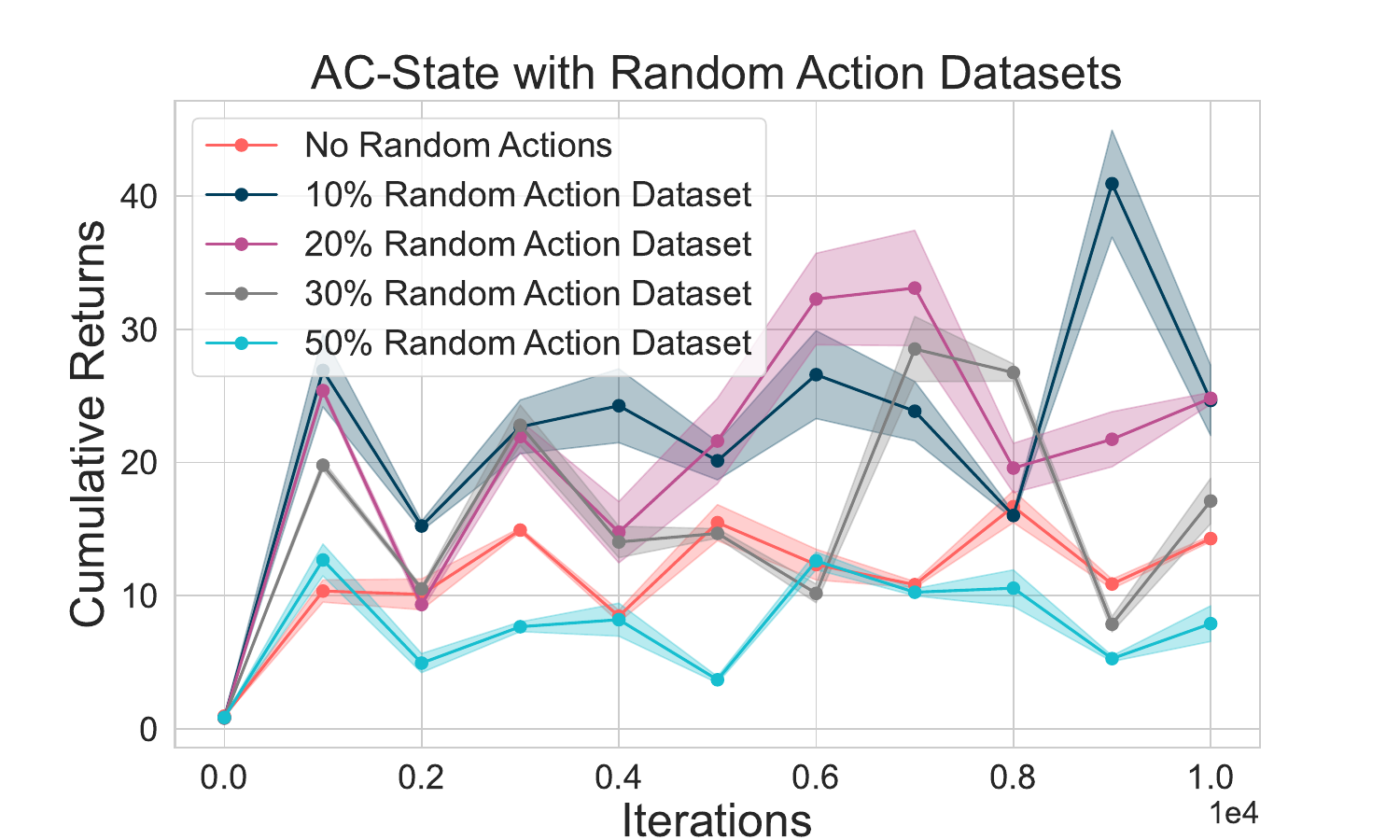}
}\\
\subfigure[DRIML \cite{mazoure2020deep}]{
\includegraphics[width=0.45\textwidth]{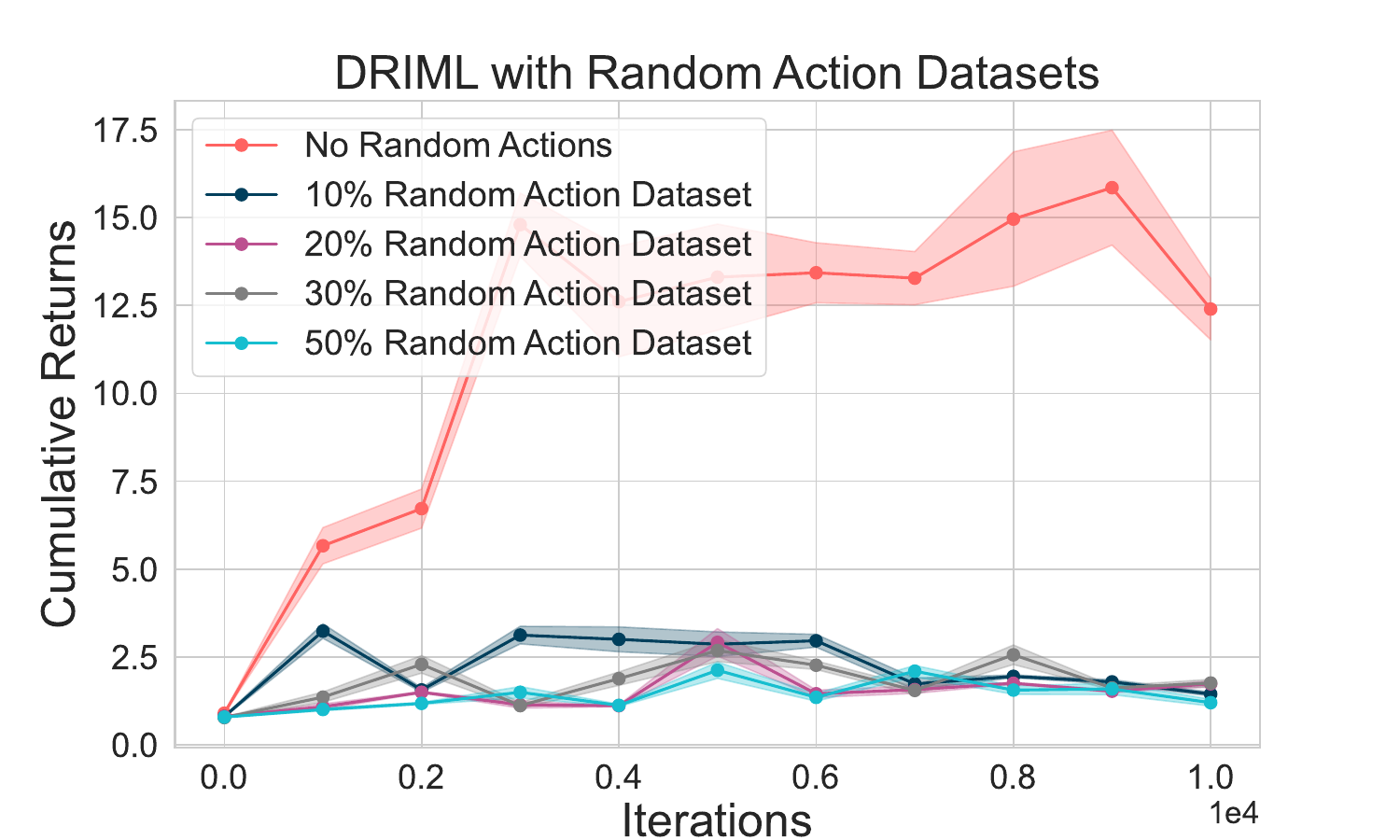}
}
\subfigure[\methodname with Sequence Prediction]{
\includegraphics[width=0.45\textwidth]{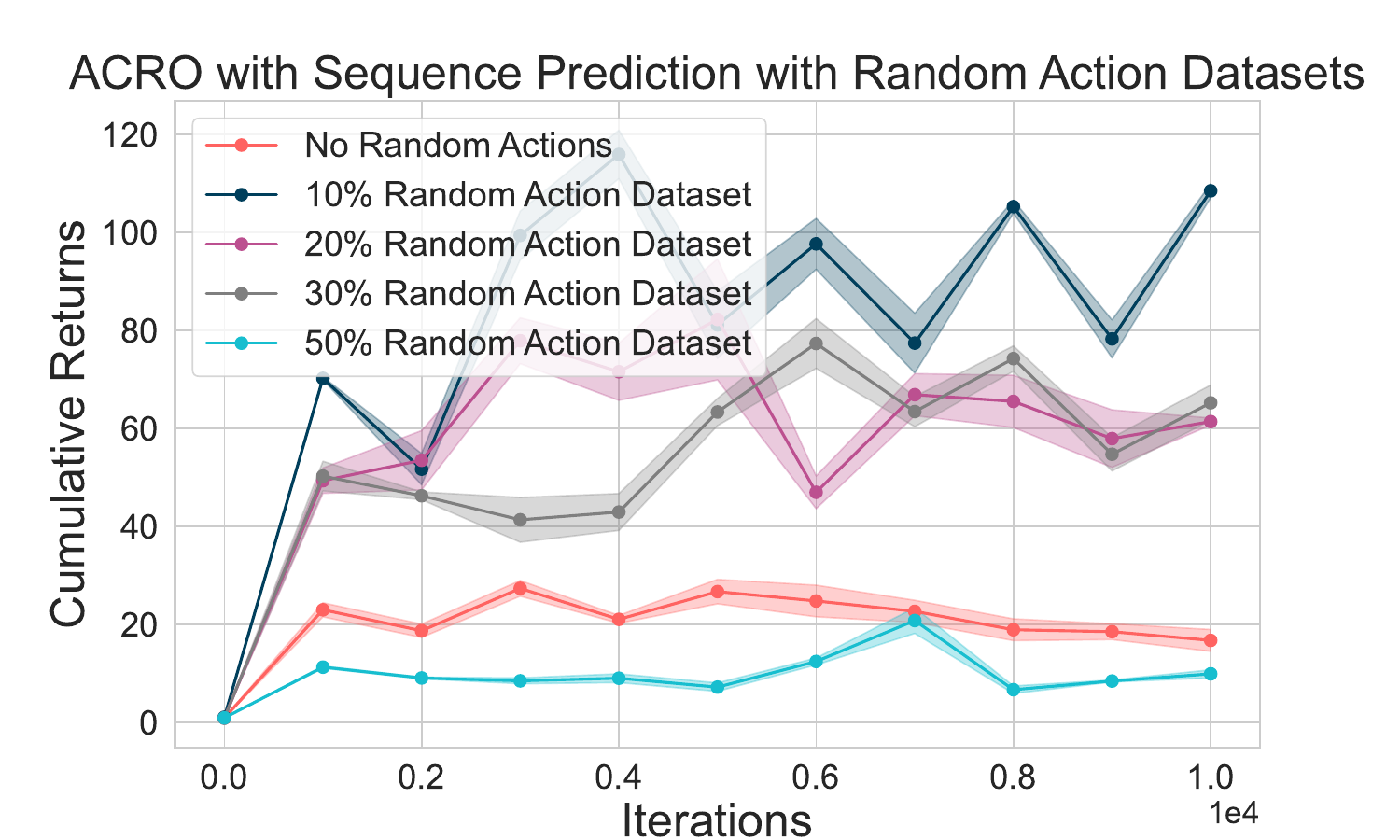}
}
\caption{\textbf{Varying amounts of coverage on the datasets}. We make the assumption that adding more random actions to the datasets means the dataset has a higher coverage of state and action space. We show that as the $\%$ of random actions on datasets increases, the performance of each method degrates, especially the ones like DRIML that are independent of action prediction. In contrast, all other methods that rely on action prediction, suffer less depending on the amount of random actions that may exist in the datasets. This shows a potential for representation objectives based on action prediction that the performance of these methods degrades least even if the dataset quality is poor (where in our case, higher coverage due to more random actions means the dataset is degrading from an expert dataset to more of a random dataset). }
\label{fig:data_coverage}
\end{figure*}

\textbf{Experiments} : We varied the amount of data coverage in the offline datasets, by taking the of times, the data collecting policy takes random actions. We vary the $\%$ of random actions from 10$\%$ to 50$\%$ where we assume that more randomness in actions taken by an expert policy means higher state space coverage. We follow the same experiment setup as before, and now show how the performance of ACRO (and two other baselines) varies as we have lower to higher coverage in the datasets, as in figure \ref{fig:data_coverage}.

\subsection{Quantitative Analysis}
\label{app:quant-analysis}
We provide results which show how ACRO representations lead to much more aligned cosine similarities as compared to DRIML and DRQ, when the background distractor is changed. We conduct a simple experiment where 1) the Exo-information is changed while keeping the Endo-information the same, and 2) the Exo-information remains the same while the Endo-information is changed (see Table~\ref{tab:quant-analysis}). For each case, we then compute the cosine similarity between representations and report normalized cosine similarities in the Table below. The normalization is w.r.t the base cosine similarity from when the Exo-information is changed while keeping the Endo-information fixed. We see that the ACRO representations consistently have higher cosine similarities, with DrQ performing second best (and even being slightly better than ACRO in one case), while DRIML representations capturing quite a bit of the background information. Note that in terms of policy returns, the ranking between methods is consistent with the cosine similarities, i.e ACRO gets the highest return, followed by DrQ and then DRIML.

\begin{table}[h!]
\caption{\textbf{Quantitative Analysis} for how cosine similarities differ when keeping exo information fixed while varying endo information and vice versa.}\label{tab:quant-analysis}
\vspace{-1mm}
\footnotesize
\tablestyle{4pt}{1.2}
\resizebox{0.4\columnwidth}{!}{%
\begin{tabular}{l|c c c}
\multicolumn{1}{l|}{\textsc{Environment}} &
  \textsc{DRIML} &
  \textsc{DrQ} &
  \textsc{ACRO} \\ 
\shline
\textbf{\hardexo b)} & 0.03 & 0.35 &\textbf{0.80} \\
\textbf{\hardexo a)} & 0.02 & 0.05 & \textbf{0.60} \\
\textbf{\mediumexo c)} & 0.11 & 0.53 & 0.45 \\
\end{tabular}
}
\end{table}

\clearpage
\section{Detailed Experiment Setup and Full Results}
\label{app:exp_setup}

We use the visual d4rl (v-d4rl) benchmark \cite{vd4rl} and additionally add various kinds of exogenous noise to visual datasets. For all our experiments, comparing \methodname with other baseline representation objectives, we pre-train the representations for $100$K pre-training steps. Given pixel based visual offline data, we use a simple CNN+MLP architecture for encoding obeservations and predicting the \methodname actions. We also use cropping-based data augmentation as in DrQv2 while pre-training the representations for all methods. Specifically, the \methodname encoder uses 4-layers of convolutions, each with a kernel size of 3 and 32 channels. The original observation is of 84 $\times$ 84 $\times$ 9, corresponding to a 3 channel-observation and a frame stacking of 3. The final encoder layer is an MLP which maps the convolutional output to a representation dimension of 256, giving the output $\phi(x)$. This is followed by a 2-layer MLP (hidden dim-256) that is used to predict the action given a 512 input corresponding to a concatenated $s_t$ and $s_{t+k}$ representations. For \methodname, we sample $k$ from 1 to 15 uniformly. We use ReLU non-linearity and ADAM for optimization all throughout.

In DrQv2, data augmentation is applied only to the images sampled from the replay buffer, and not during the sample collection procedure. Given the pixel based control tasks, where the images are $84 \times 84$, DrQ pads each side by $4$ pixels (repeating boundary pixels) and then selects a random $84 \times 84$ crop, yielding the original image, shifted by $±4$ pixels. This procedure is repeated every time an image is sampled from the replay buffer; and makes DrQ data augmentation quite effective based on the random shifts alone, without the need for any additional auxiliary losses. We now provide a detailed summary of the different types of exogenous information based datasets as demonstrated in \pref{fig:illustration_experiment_setups}. 

\subsection{Easy-Exogenous Information (\easyexo)}

The visual RL offline benchmark proposed in \citep{vd4rl} provided pixel-based offline datasets collected using varying degrees of a soft actor-critic (SAC) policy. Furthermore, \citep{vd4rl} proposed a suite of distractor based datasets with different levels of severity in distractor shift, ranging from easy-shift, medium-shift to hard-shift. In the \easyexo setting, we first consider pixel-based offline data without and with visual distractors, as shown in~\pref{tab:easy_exo_results}. Experimental results in~\pref{tab:easy_exo_results} show that even without any exogenous noise, $\methodname$ learns agent-centric latent state representations more accurately than the different state-of-the-art baselines, such that by efficiently decoupling the endogenous state from the exogenous states, policy learning during the offline RL algorithm can lead to significantly better evaluation performance compared to several other baselines. 

\begin{figure*}[t]
\centering
{
\includegraphics[width=0.7\textwidth]{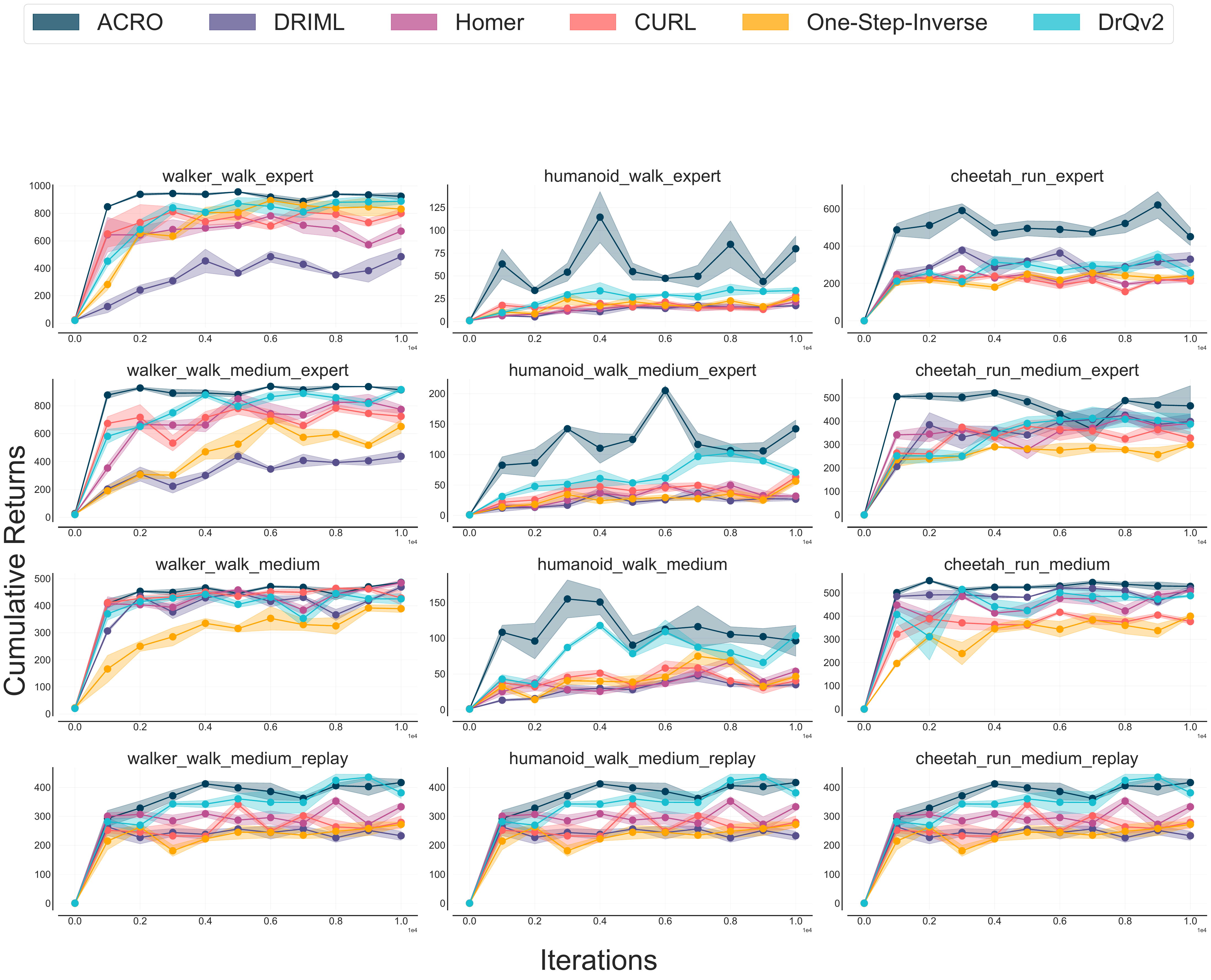}
}
\caption{\textbf{\easyexo-No Distractors Full Results}.}
\label{fig:vd4rl_main_pre_train}
\end{figure*}

\textbf{Visual Offline Control (V-D4RL) without Distractors}. We first verify the effectiveness of learning representations with $\methodname$ without any additional exogenous distractors, and compare with several baselines for learning representations. \pref{fig:vd4rl_main_pre_train} provides detailed performance curves for \pref{tab:easy_exo_results}. 

\begin{figure}[t]
    \centering
    \includegraphics[width=0.4\textwidth]{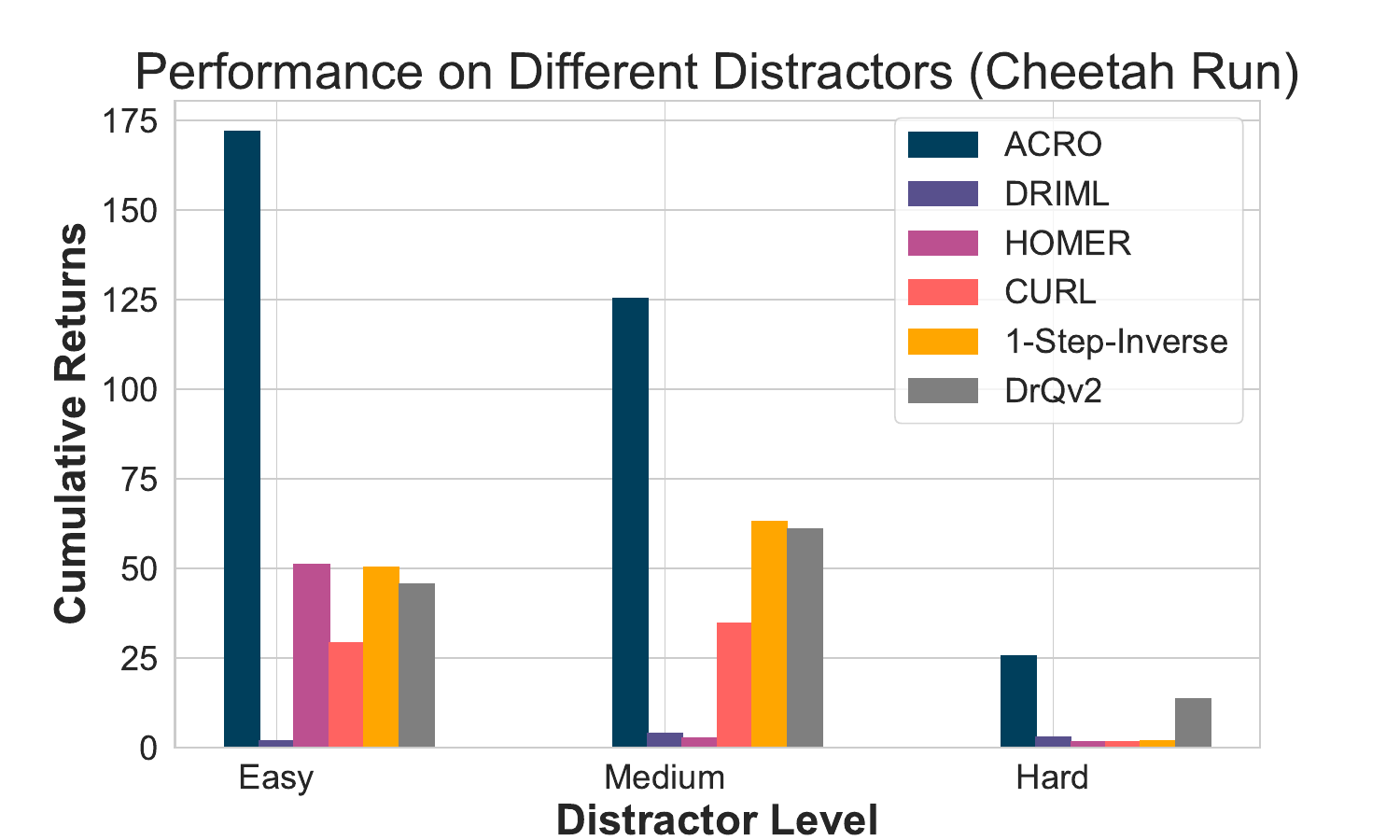}
    \includegraphics[width=0.4\textwidth]{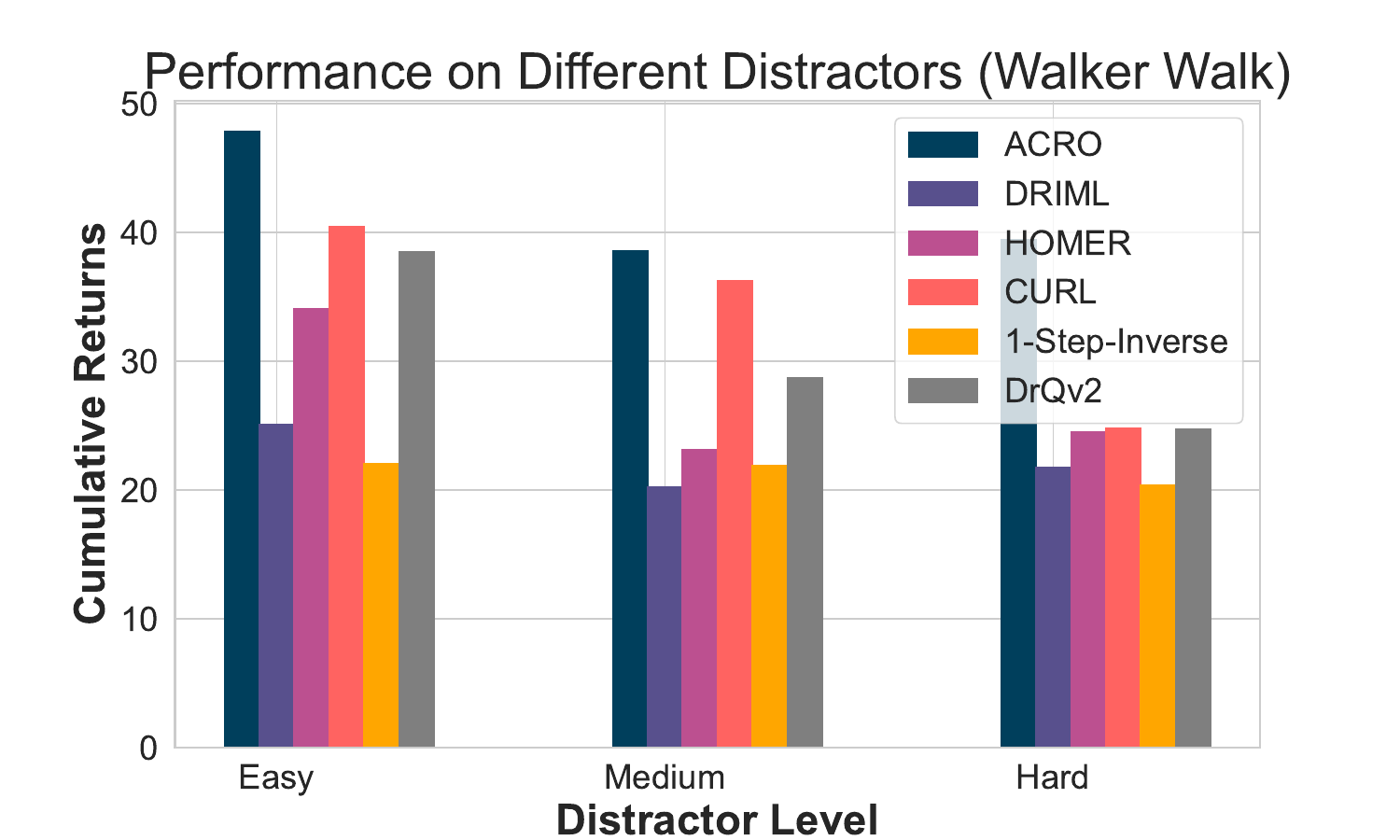}\\
    \caption{\textbf{\easyexo-Image Distractors Full Results}.}
    \label{fig:vd4rl_distractor_barplots}
\end{figure}

\textbf{V-D4RL with Varying Severity of Distractor Data Shift}. We then consider the distractor setting in v-d4rl benchmark \citep{vd4rl} with varying levels of distractor difficulty. Here, the exogenous noise is based on background static image distractors inducing a distribution shift in the dataset, depending on the level of difficulty from \textit{easy}, to \textit{medium} to \textit{hard} distractors. As shown in \pref{fig:vd4rl_distractor_barplots}, we consider two different domains and find that with varying difficulty levels, $\methodname$ can consistently outperform several state of the art baselines, learning directly from pixel data.

\subsection{Medium-Exogenous Information (\mediumexo)}

\begin{figure*}[h]
\centering
\centering
{
\includegraphics[width=0.7\textwidth]{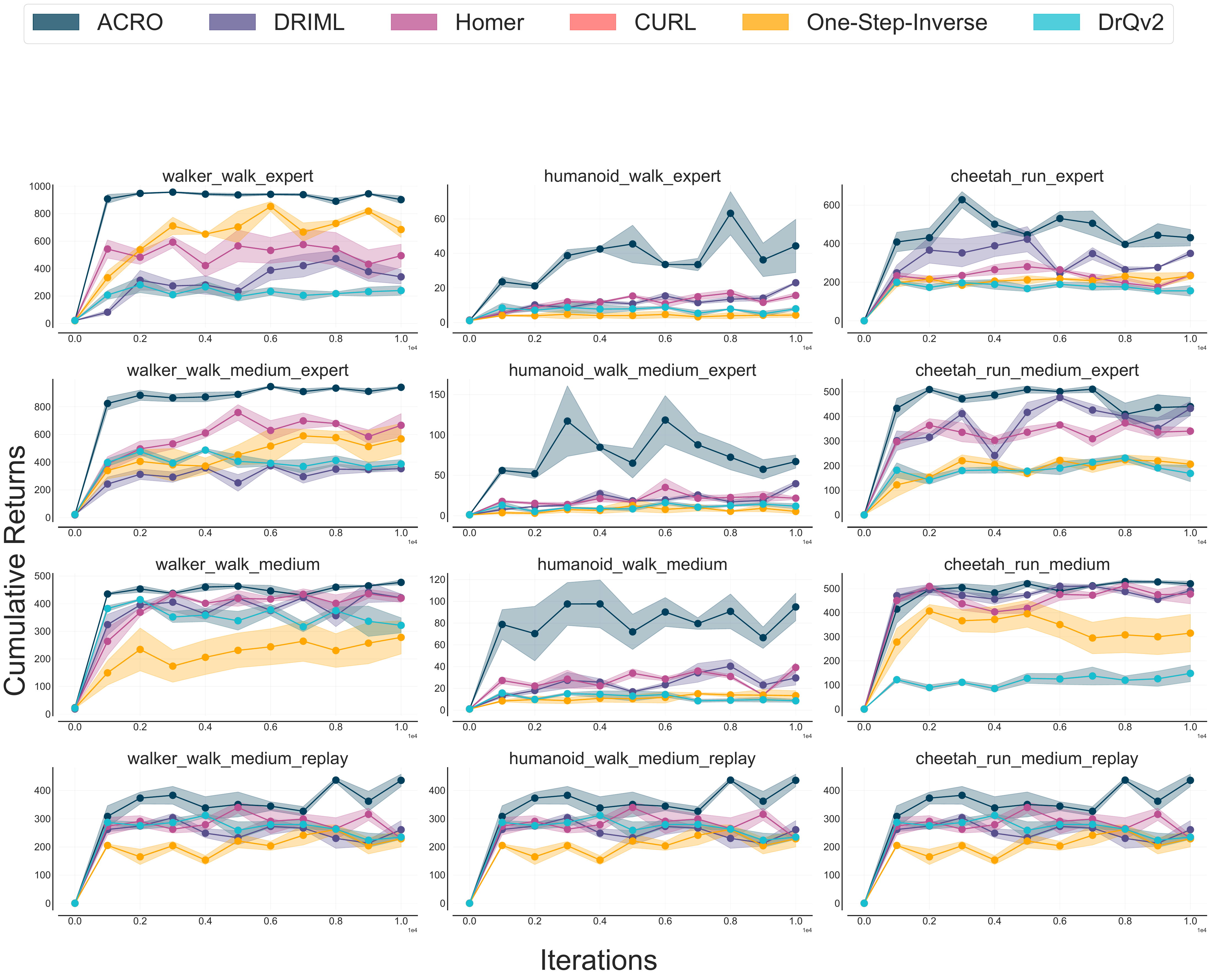}
}
\caption{\textbf{\mediumexo-Corner Full Results}.}
\label{fig:vd4rl_corner_distractor}
\end{figure*}

We then consider three different types of medium exogenous information that might appear in visual offline data. To that end, we consider exogenous uncorrelated images from STL-10 image dataset \citep{stl10} that appear on the corner or the side of the agent observation, and the goal of $\methodname$ is to filter out the exogenous information while recovering only the agent state. We consider \textit{three different types of \mediumexo information:},

\textbf{Exogenous Image Distractors Placed on the Corner of Agent Observation}. The exogenous image from STL10 dataset appears in the corner of the agent observation. This does not change the observation size of the agent; and we simply add the exogenous image in one corner, which is fixed during an entire episode during the offline data collection.~\pref{fig:vd4rl_corner_distractor} summarizes the result with different exogenous images placed in the corner. $\methodname$ consistently outperforms several other baselines for a range of different datasets, since it can suitably filter out the exogenous part of the agent state.

\begin{figure*}[h]
\centering
\centering
{
\includegraphics[width=0.7\textwidth]{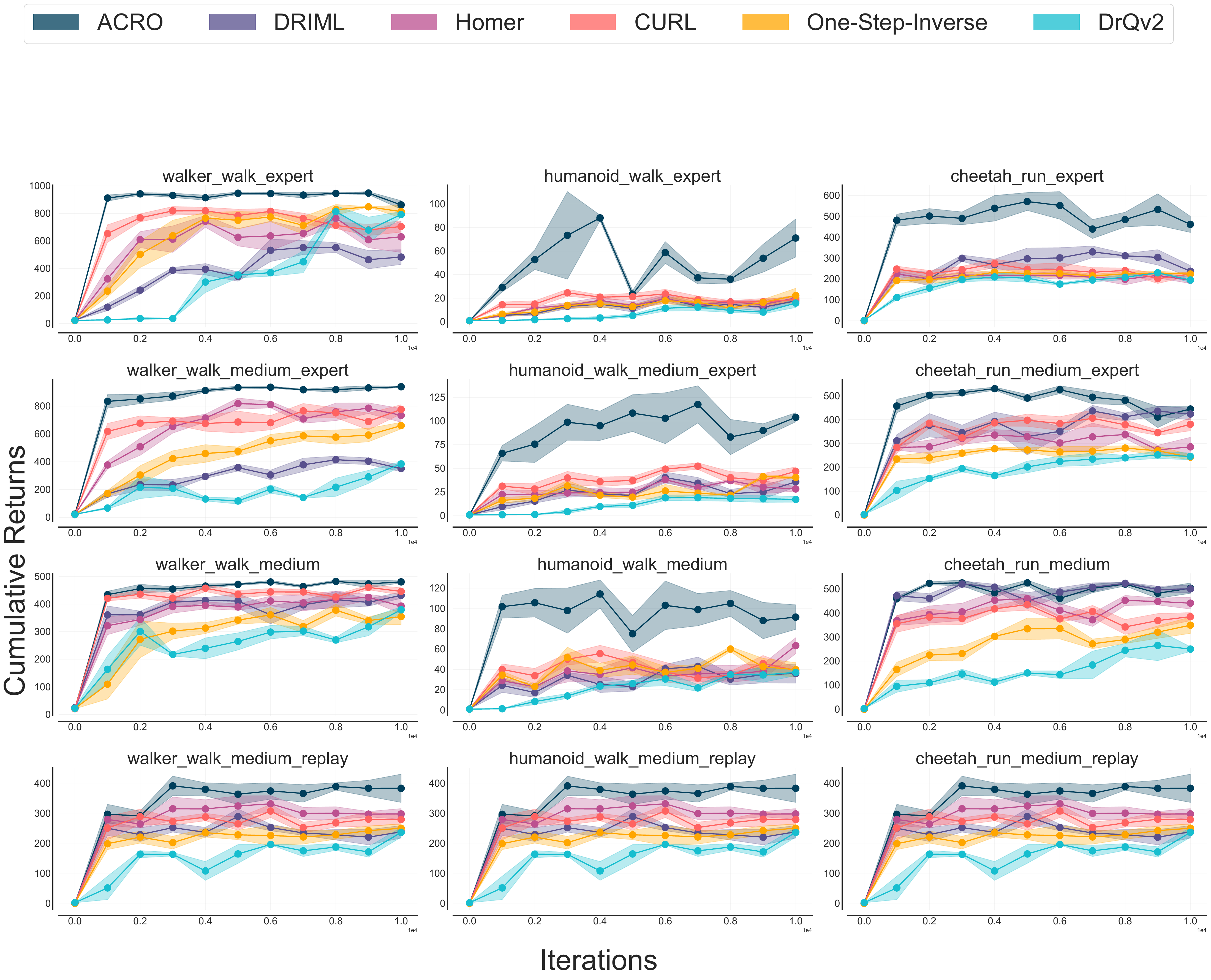}
}
\caption{\textbf{\mediumexo-Side Full Results}.}
\label{fig:vd4rl_side_distractor}
\end{figure*}

\textbf{Exogenous Image Distractors Placed on the Side of Agent Observation}. A slightly more difficult setting where now the STL10 exogenous image appears on the side of the agent observation space. This augments the agent observation space from 84 $\times$ 84 $\times$ 3 to  84 $\times$ $84$ $\times$ 2 $\times$ 3 since we consider downsampled STL10 images.~\pref{fig:vd4rl_side_distractor} summarizes this result comparing $\methodname$ with the baseline representation objectives. 

\begin{figure*}[h]
\centering
{
\includegraphics[width=0.7\textwidth]{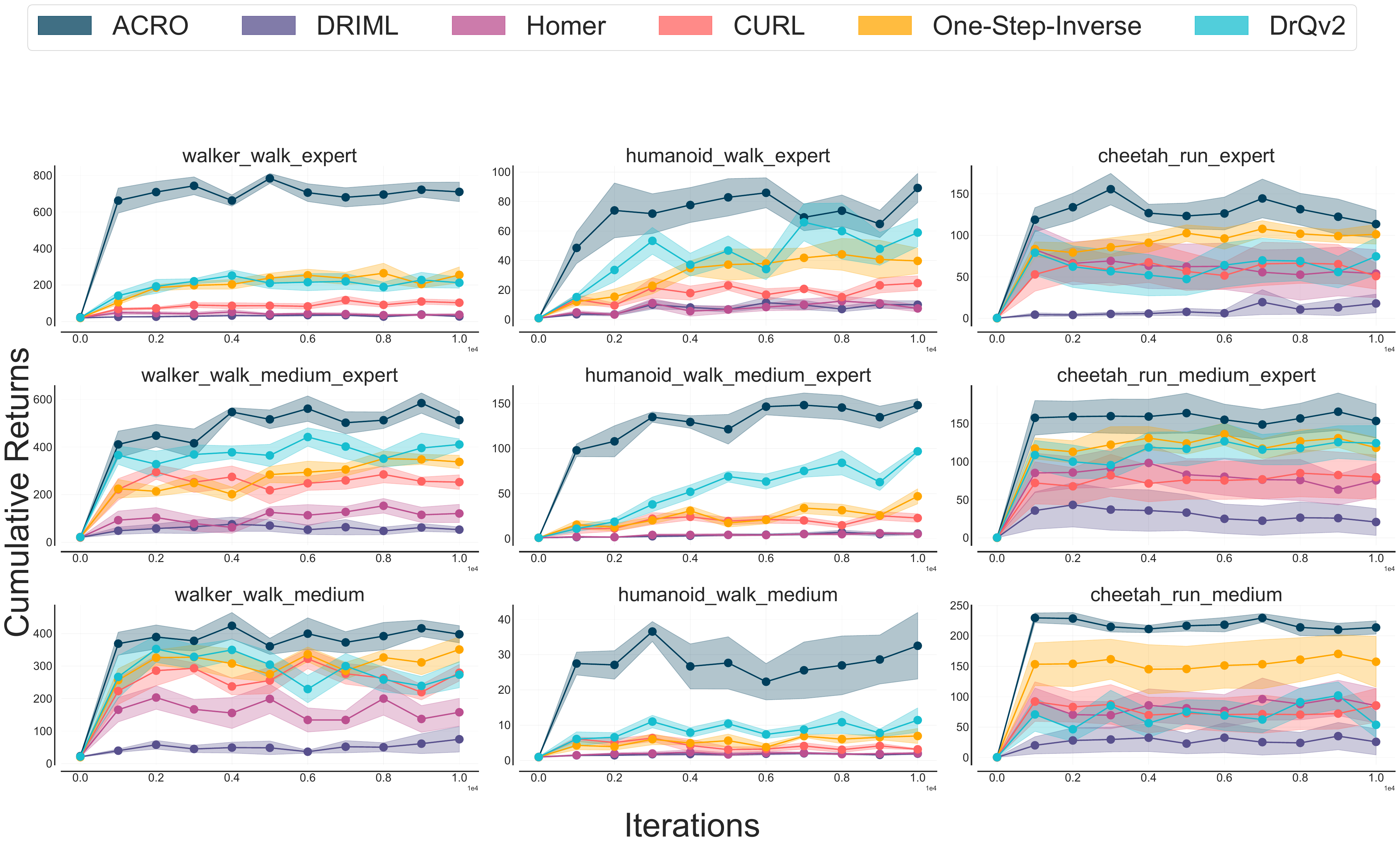}
}
\caption{\textbf{\mediumexo-Fixed Video Full Results}.}
\label{fig:fixed_video_distractor}
\end{figure*}

\textbf{Fixed Video Distractor}. Finally we consider the distractor setting that has appeared in prior works in online RL \citep{DBC} with fixed video distractors playing in the background of agent observation space. For this setting, we specifically re-collect the dataset following the procedure in \citep{vd4rl} where the SAC data collecting agent also sees a fixed video distractor playing in the background.~\pref{fig:fixed_video_distractor} summarizes the result and shows performance plots where $\methodname$ can significantly outperform all the baselines across all different types of exogenous datasets. 

\subsection{Hard Exogenous Information (\hardexo)}

We finally consider three sets of different hard exogenous information settings, and find that these $\hardexo$ can remarkably make it difficult for existing state of the art representation objectives to learn underlying agent-centric states. This setting provides evidence that under suitably constructed exogenous information, which appear highly time correlated during the offline data collection, the baseline methods can fail to capture underlying agent-centric latent states. In contrast, the objective we consider in $\methodname$, along with the theoretical guarantees for learning a suitable encoder to recover the endogenous states, shows that policy optimization based on the agent-centric latent states can lead to efficient learning in these control tasks. We consider \textit{three different types of \hardexo information:} 

\begin{figure*}[h]
\centering
{
\includegraphics[width=0.7\textwidth]{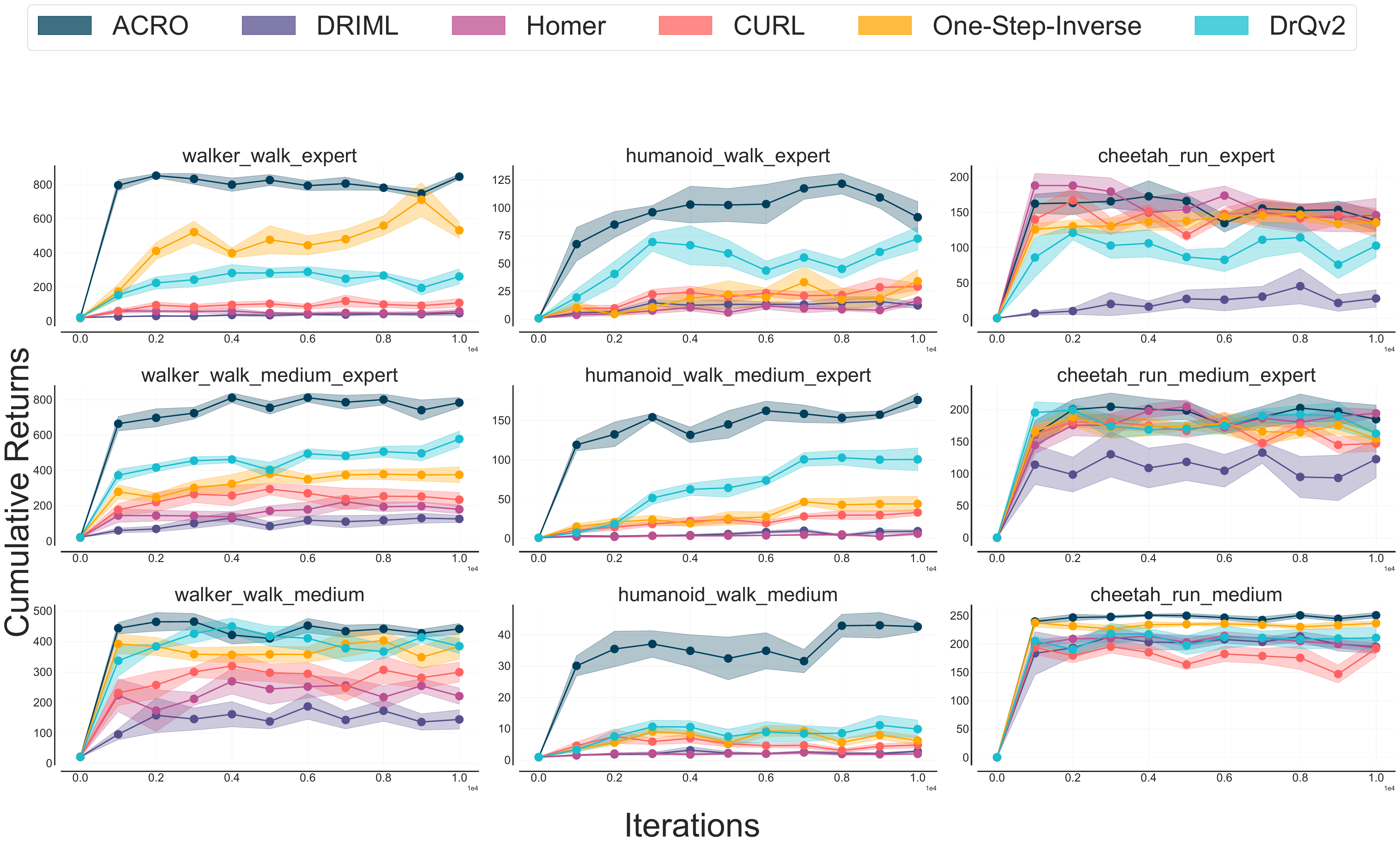}
}
\caption{\textbf{\hardexo-Static Image Full Results}.}
\label{fig:correlated_background}
\end{figure*}

\textbf{Static Background Image that Changes Per Episode}. We first consider time correlated static images appearing in the background of the agent observation. For this setting, during data collection, the agent sees a fixed image in the background for an entire episode, and it changes per episode of data collection. This time correlated exogenous information ensures that the baselines can remarkably get distracted, while $\methodname$ can still be robust to the static image background.~\pref{fig:correlated_background} summarizes the results and shows that several existing representation baselines can fail due to time correlated static image distractors. 

\begin{figure*}[h]
\centering
\centering
{
\includegraphics[width=0.7\textwidth]{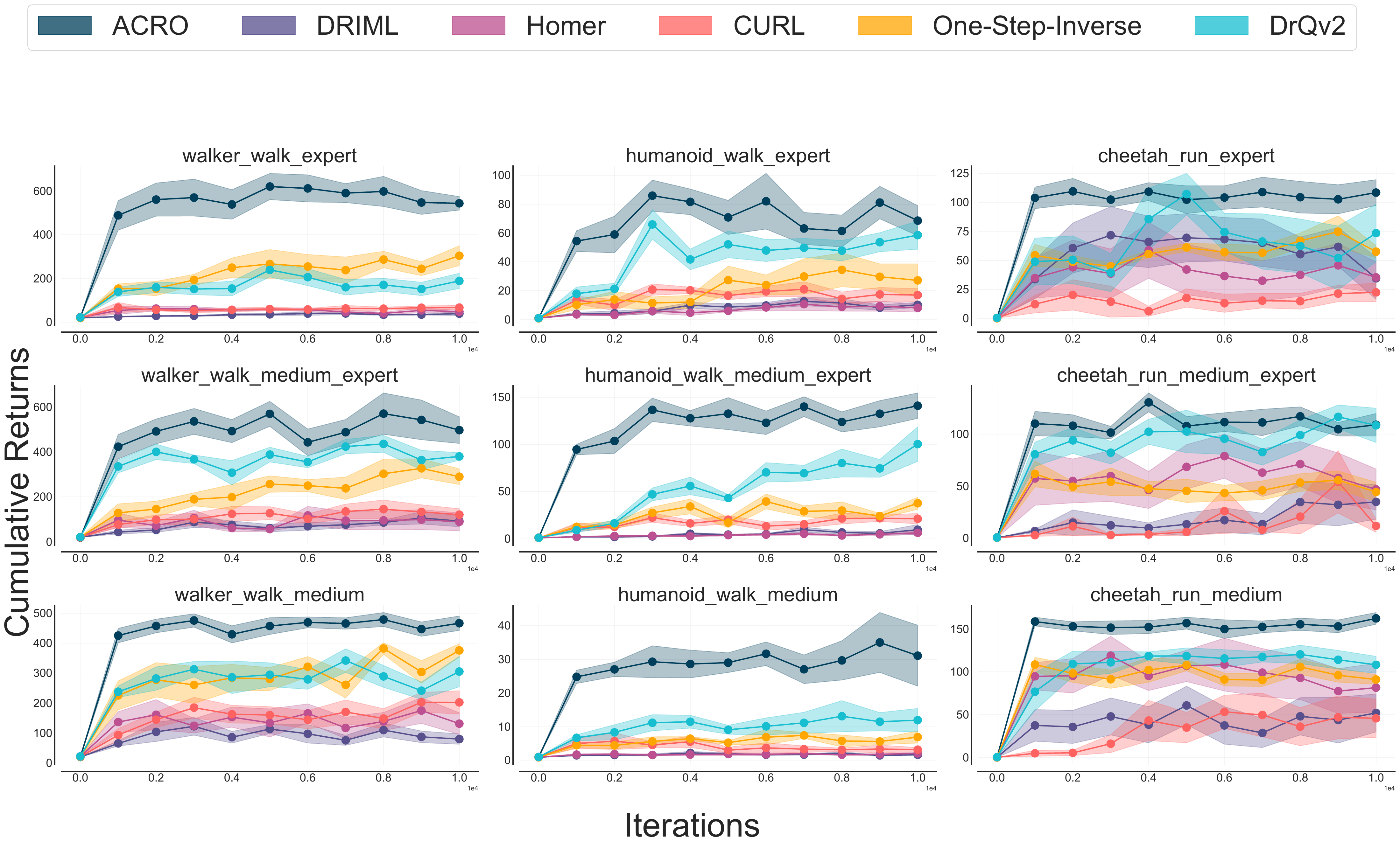}
}
\caption{\textbf{\hardexo-Video Full Results}.}
\label{fig:changing_video_distractor}
\end{figure*}

\textbf{Exogenous Video Distractors that Changes Per Episode}. We then consider an even more difficult \hardexo setting where now the video distractors playing in background also changes per episode during data collection. This is a novel setting with video distractors in RL, since we explicitly consider diverse set of background videos which also changes per episode of data collection. Similar to the above,~\pref{fig:changing_video_distractor} summarizing the results with diverse video distractors in background per episode, shows that this setting can also break the baseline representation learners to recover the agent-centric latent states, while performance of $\methodname$ remains robust to it, since the $\methodname$ objective learns encoder to recover the endogenous agent-centric latent states accurately. 

\begin{figure*}[h]
\centering
\centering
{
\includegraphics[width=0.7\textwidth]{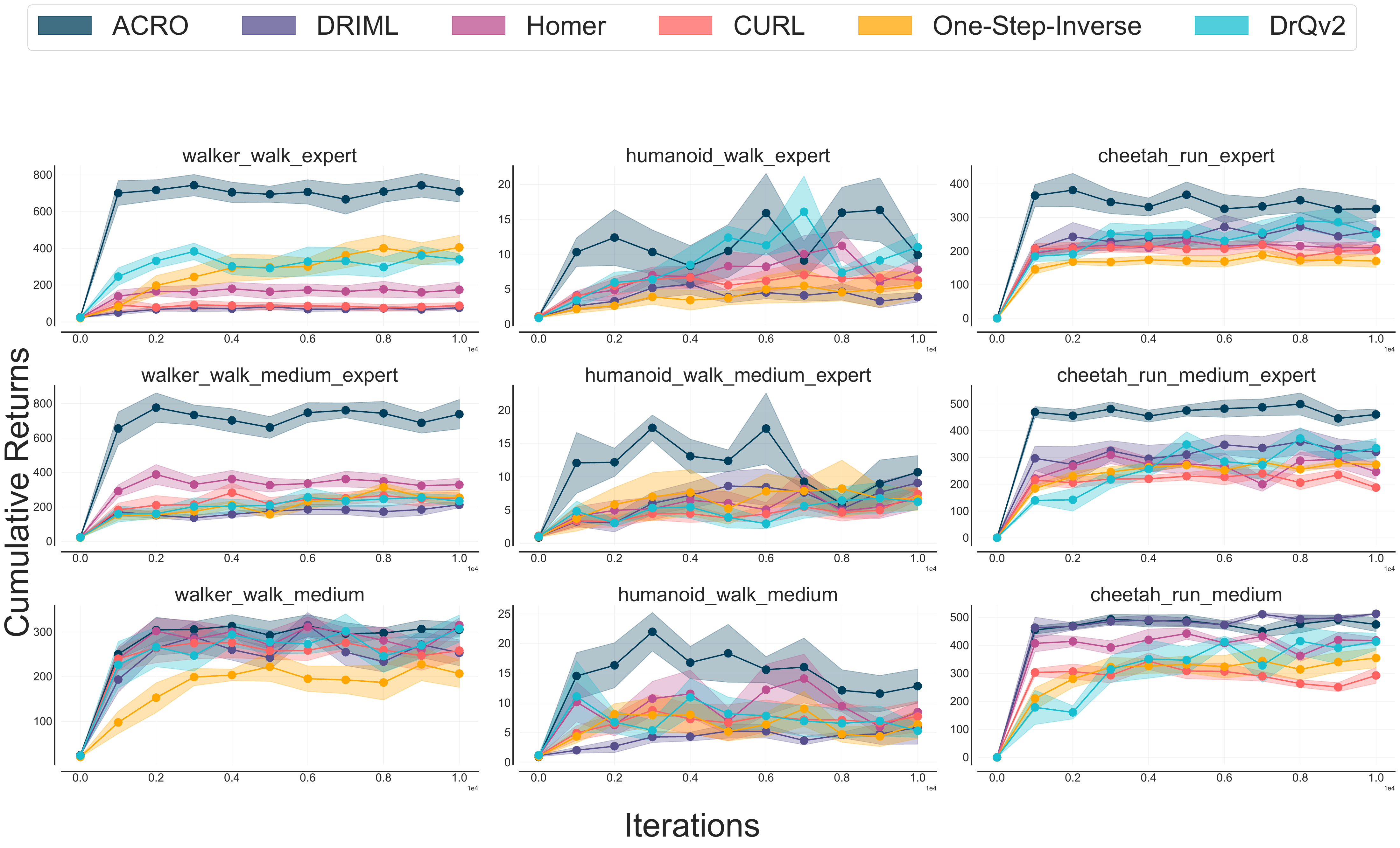}
}
\vspace{-2mm}
\caption{\textbf{\hardexo-Background Agent Full Results}.}
\label{fig:multi_maze_exo_grid_full}
\end{figure*}

\textbf{Multi-Environment Agent Observations as Exogenous Information}. Finally, we consider the most challenging \hardexo where now in addition to the environment observation, the agent additionally sees other random action agent observations. Here, the goal of the agent is to learn representations to identify the \textit{controllable} environment, while other random-action observations are \textit{uncontrollable} or exogenous to the agent. This is quite a difficult task since the agent we are tring to control also sees observations from the same domain, of other agents playing with random actions. The agent-centric agent observation now consists of other agents placed in a 3 $\times$ 3 grid.~\pref{fig:multi_maze_exo_grid_full} summarizes the experiment results  showing that $\methodname$ significantly outperforms all baseline representation learners. 

\end{document}

%% file: math_commands.tex
\usepackage{amsmath,amsfonts,bm}

\DeclareMathOperator*{\E}{\mathbb{E}}

\def\eqref#1{equation~\ref{#1}}

\def\1{\bm{1}}

\def\rva{{\mathbf{a}}}

\def\rvx{{\mathbf{x}}}

\DeclareMathAlphabet{\mathsfit}{\encodingdefault}{\sfdefault}{m}{sl}
\SetMathAlphabet{\mathsfit}{bold}{\encodingdefault}{\sfdefault}{bx}{n}

\DeclareMathOperator*{\argmax}{arg\,max}

\def\rva{{\mathbf{a}}}

\def\rvx{{\mathbf{x}}}

\DeclareMathAlphabet{\mathsfit}{\encodingdefault}{\sfdefault}{m}{sl}
\SetMathAlphabet{\mathsfit}{bold}{\encodingdefault}{\sfdefault}{bx}{n}

